\documentclass[11pt]{article}

\usepackage[final]{acl}

\usepackage{times}
\usepackage{latexsym}

\usepackage[T1]{fontenc}

\usepackage[utf8]{inputenc}

\usepackage{microtype}

\usepackage{inconsolata}

\usepackage{graphicx}

\usepackage{mathrsfs}
\usepackage{tikz}
\usepackage{amsmath}
\usepackage{graphicx}
\usepackage{pgffor}
\usepackage{float}
\usepackage{booktabs}
\usepackage{colortbl}
\usepackage{tabularx}
\usepackage[table]{xcolor}
\usepackage{pbox}
\usepackage{amsfonts, amssymb}
\usepackage{pifont}
\usepackage{cleveref}
\usepackage{hyperref}
\usepackage{makecell}
\usepackage{verbatim}
\usepackage{paralist}
\usepackage[usestackEOL]{stackengine}
\PassOptionsToPackage{table}{xcolor}
\usepackage{mathtools, nccmath}
\usepackage{arydshln}

\usepackage{lipsum}

\usepackage{enumitem}

\usepackage[ruled,vlined]{algorithm2e}
\usepackage{algpseudocode}

\usepackage{xcolor}
\usepackage{xstring}
\usepackage{siunitx}
\usepackage{subcaption}
\usepackage{multirow}
\usepackage{rotating}

\usepackage{nicematrix}

\usepackage{amsthm}

\newtheorem{proposition}{Proposition}[]
\newtheorem{property}{Property}[]
\newtheorem{corollary}{Corollary}[]
\newtheorem{lemma}{Lemma}[]

\newcommand{\huggingfacedown}{\includegraphics[height=0.75em]{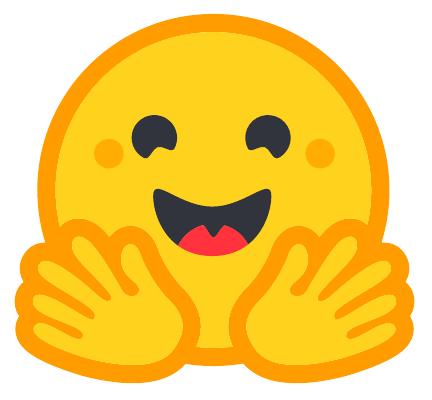}}
\newcommand{\githubdown}{\includegraphics[height=0.75em]{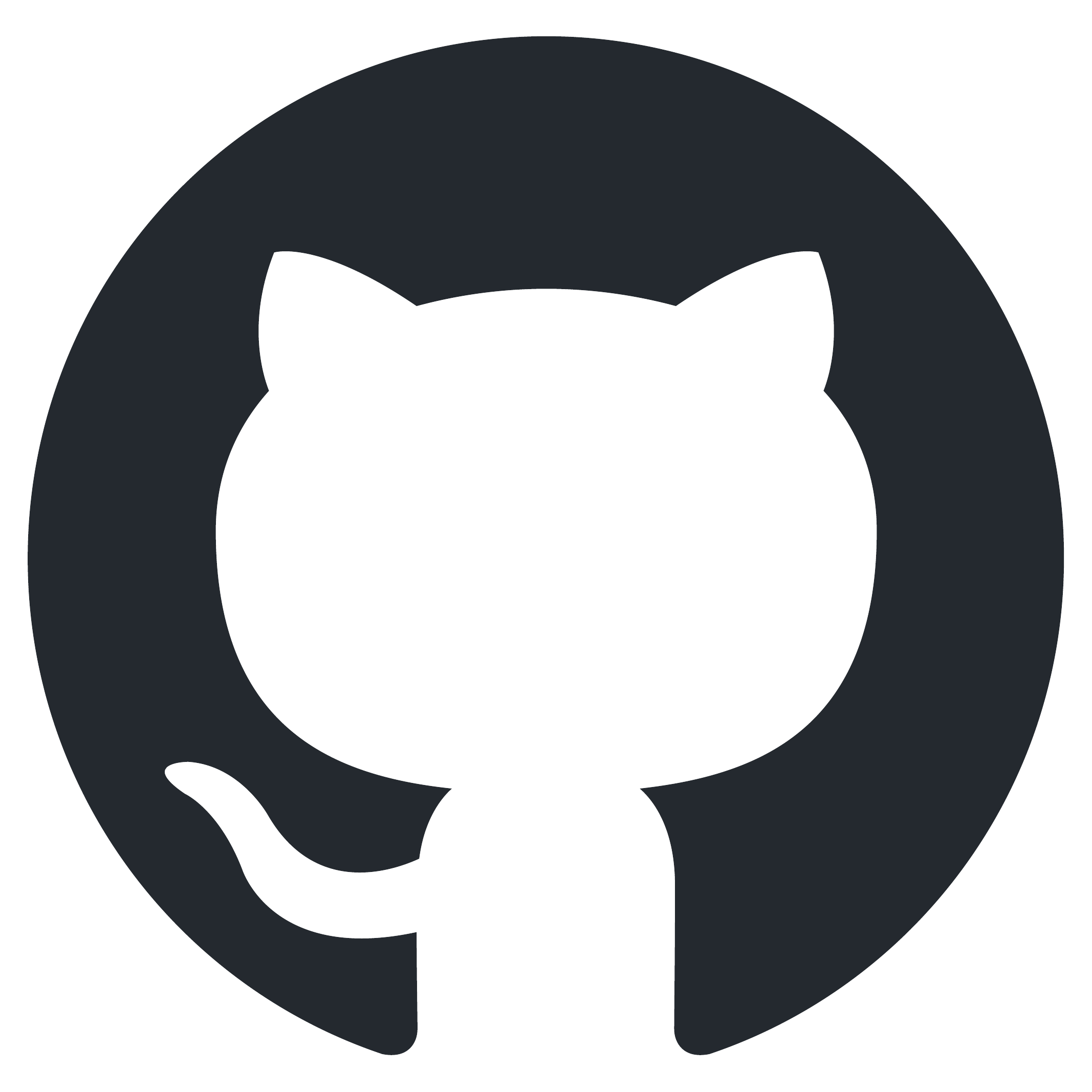}}

\usepackage{tablefootnote}

\title{MultiHashFormer: Hash-based Generative Language Models}

\author{
Huiyin Xue$^{\mu}$\thanks{Work done while at the University of Sheffield.} \quad Atsuki Yamaguchi$^\sigma$ \quad Nikolaos Aletras$^\sigma$\\
$^\mu$Mohamed bin Zayed University of Artificial Intelligence, United Arab Emirates\\
$^\sigma$University of Sheffield, United Kingdom\\
\texttt{huiyin.xue@mbzuai.ac.ae} \quad \texttt{\{ayamaguchi1, n.aletras\}@sheffield.ac.uk}
}

\begin{document}

\newcommand{\colorednum}[1]{%
    \ifnum\fpeval{#1 > 0} = 1
        \textcolor{teal!80!black}{\small $\uparrow$ \num[minimum-decimal-digits=2, round-mode=places, round-precision=2]{\fpeval{round(#1,2)}}}%
    \else
        \ifnum\fpeval{#1 < 0} = 1
            \textcolor{red!70!black}{\small $\downarrow$ \num[minimum-decimal-digits=2, round-mode=places, round-precision=2]{\fpeval{round(abs(#1),2)}}}%
        \else
            \small #1
        \fi
    \fi
}

\maketitle

\begin{abstract}
Language models (LMs) represent tokens using embedding matrices that scale linearly with the vocabulary size. To constrain the parameter footprint, prior work proposes hashing many tokens into a single vector within encoder-only models. While this offers parameter efficiency, many-to-one collisions prevent its use in causal LMs. In this paper, we propose \textsc{MultiHashFormer}, a new framework that allows hash-based autoregression. Each token is represented as a unique hash signature, a short sequence of discrete hash IDs, generated by multiple independent hash functions. A Hash Encoder compresses this signature into a single latent vector for processing by a Transformer decoder. Then, a Hash Decoder generates the hash signature of the next token, which is subsequently mapped back to text. We evaluate our approach at the 100M, 1B and 3B parameter scales, demonstrating that \textsc{MultiHashFormer} consistently outperforms standard Transformer LMs across multiple benchmarks. Furthermore, we show that our model handles multilingual vocabulary expansion with a constant parameter footprint 
without any modifications.\looseness=-1

\githubdown\quad\url{https://github.com/HUIYINXUE/MHF}\\
\huggingfacedown\quad\url{https://hf.co/papers/2606.28057}

\end{abstract}

\section{Introduction} \label{sec:introduction}

Language models~\citep[LMs]{olmo2026olmo3,yang2025qwen3technicalreport,abdin2024phi4technicalreport} typically map tokens into high-dimensional vectors via learned embedding matrices. This allows each token in a discrete vocabulary to be represented by a unique, dense vector. However, this linear scaling creates a \textit{vocabulary bottleneck}, locking the model into a fixed token capacity and restricting its ability to adapt seamlessly to new domains, or languages.

To mitigate this, previous research has explored token hashing~\citep{prakash-etal-2020-compressing,shu2018compressing,svenstrup2017hash,ganchev-dredze-2008-small} to fix the parameter footprint of the embedding space. Hash-based models such as Proformer~\citep{sankar-etal-2021-proformer} and HashFormer~\citep{xue-aletras-2022-hashformers} use many-to-one mappings, where multiple tokens (e.g., \textit{cat}, \textit{map}, and \textit{physics}) may share the same hash index (e.g., 40). While parameter-efficient, these models are restricted to encoder-based architectures and discriminative training. In generative settings, if a decoder predicts a shared hash index, the model cannot deterministically recover the intended token from the set of colliding candidates. Consequently, this ambiguity has made training causal LMs on hashed spaces practically infeasible.

\begin{figure*}[!t]
    \centering
    \includegraphics[width=\linewidth]{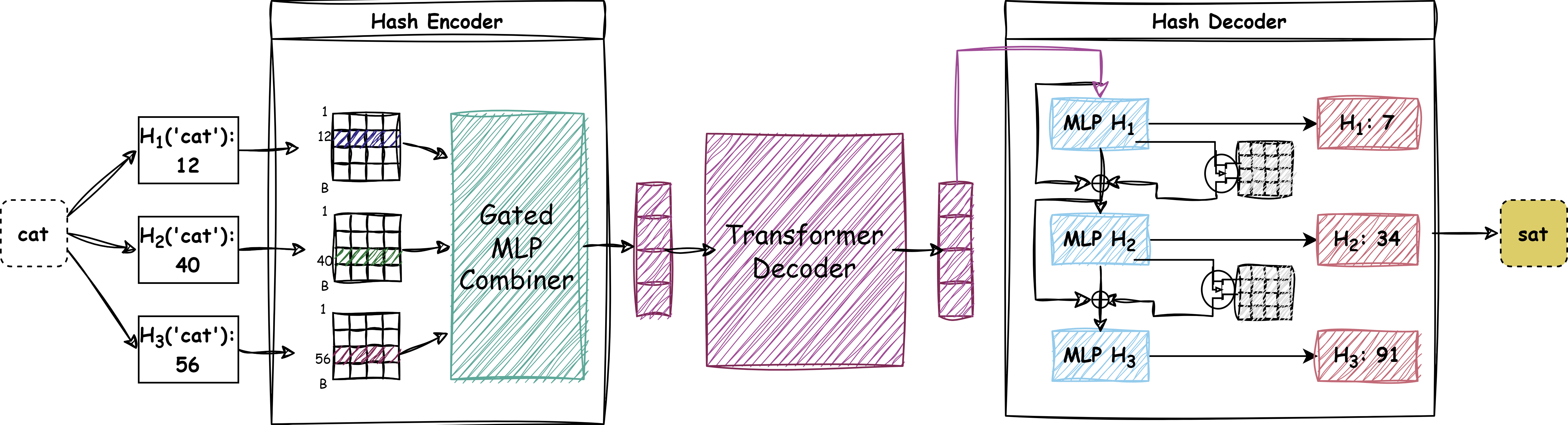}
    \caption{Overview of the \textsc{MultiHashFormer} framework using three different hash functions (for illustration purposes) with $B$ buckets to obtain a multi-ID hash signature.}
    \label{fig:overview}
\end{figure*}

In this paper, we propose \textsc{MultiHashFormer}, a new  framework designed to bypass the vocabulary bottleneck for \textit{decoder-based LMs}. Drawing inspiration from chaotic dynamic memory systems~\cite{skarda1987brains} which leverage distributed, overlapping state spaces for high-capacity pattern retrieval~\citep{bricken2021attention}, we replace the traditional embedding matrix with a modular \textit{hashing interface}. 
Specifically, each token is converted into a \textit{unique hash signature} consisting of a short sequence of discrete hash IDs generated by multiple independent hash functions, similar to chaotic transitions that allow biological networks to chain together vast vocabularies of concepts~\cite{tsuda2001toward}. For example, \textit{cat} is  represented as $[12, 40, 56]$ and \textit{map} as $[99, 40, 3]$. This multi-ID mapping eliminates the token collision problem of previous hash-based methods while allowing the model to \textit{scale its vocabulary capacity with a strictly sub-linear parameter footprint}. For example, using four hash functions and 16,000 buckets per function, \textsc{MultiHashFormer} theoretically supports an upper bound of $16000^{4}$ (approx. 65 quadrillion) unique signatures.

The interface is independent of the sequence processing backbone (e.g., a Transformer~\citep{vaswani2017attention}) and manages the bidirectional translation between discrete tokens and the hash signatures. At the input, a \textit{gated compositional embedding} compresses the multi-ID signature into a dense latent vector, providing the backbone with a unified representation for sequence processing (\textit{Hash Encoder}). At the output, a \textit{cascaded predictor} reconstructs the hash signature of the next token sequentially, which is then deterministically mapped back to a specific discrete text token in the vocabulary (\textit{Hash Decoder}). Figure~\ref{fig:overview} illustrates the \textsc{MultiHashFormer} architecture.

Our main contributions are as follows:
\begin{itemize}
    \item We introduce the first hash-based framework that supports causal language modeling by preventing token collisions through multi-ID signature generation.
    \item \textsc{MultiHashFormer} models consistently outperform standard Transformer LMs at 100M, 1B and 3B scales across 10 tasks, while offering better rare word representations.
    \item We show that our models maintain performance while expanding the vocabulary size from 32K to 48K without any structural changes, or parameter count increase.
\end{itemize}

\section{Related Work}

\subsection{Token-Free Models}\label{sec:token-free}
LMs rely on an embedding matrix that scales linearly with the vocabulary size. While subword tokenizers like BPE \cite{sennrich-etal-2016-neural} and SentencePiece \cite{kudo-richardson-2018-sentencepiece} mitigate vocabulary explosion, they remain constrained by a fixed, data-derived vocabulary that struggles with rare or out-of-domain words.

Token-free models such as AU-Net~\citep{videau2026bytes}, Bolmo~\citep{minixhofer2025bolmo}, H-Net~\citep{hwang2025dynamic}, BLT~\cite{pagnoni-etal-2025-byte}, CANINE \cite{clark-etal-2022-canine}, and ByT5 \cite{xue-etal-2022-byt5} bypass the vocabulary bottleneck entirely by operating directly on unicode or byte sequences. However, this approach dramatically increases the input sequence length.
Alternatively, T-FREE \cite{deiseroth-etal-2024-free} avoids tokenization by embedding words via sparse activations over locality-sensitive hashed character trigrams. While parameter-efficient, it  relies on character-level morphological similarity, making it language-dependent. 
In contrast, \textsc{MultiHashFormer} is orthogonal to these approaches. It resolves the vocabulary bottleneck while retaining the sequence compression advantages of subword tokenization.
By decoupling the parameter matrix from the discrete vocabulary using combinatorial multi-hash mapping rather than sub-character heuristics, \textsc{MultiHashFormer} remains language-agnostic and supports any arbitrary tokenization strategy.

\subsection{Factorized and Hash-Based Embeddings}
To compress the memory footprint of standard embedding matrices, prior work has explored parameter reduction techniques. For example, ALBERT \cite{lan2019albert} uses matrix factorization to decouple the embedding dimension from the hidden dimension. While this reduces the parameter footprint, it still allocates a localized, explicit vector per token, failing to break the linear scaling constraint. 

A different approach is to use random hash embeddings \cite{svenstrup2017hash}, which map discrete tokens into a highly compressed set of physical buckets.
Architectural extensions like Proformer \cite{sankar-etal-2021-proformer} and HashFormer \cite{xue-aletras-2022-hashformers} demonstrate that token hashing can be used to train Transformer-based models.
However, compressing a vast vocabulary into a restricted physical bucket space inevitably forces multiple tokens to share the same index. This collision on a single hash function level prevents deterministic token recovery during autoregressive generation, restricting these methods strictly to encoder-only architectures.
\textsc{MultiHashFormer} resolves this limitation via the multi-identifier framework detailed in \S\ref{sec:method}.

\section{MultiHashFormer} \label{sec:method}

\textsc{MultiHashFormer} comprises three modules shown in Figure~\ref{fig:overview}: (1) a \textbf{Hash Encoder} that maps a discrete input token to a distributed multi-ID signature and compresses this signature into a single dense embedding; (2) a \textbf{Sequence Processing Backbone} that converts these embeddings into contextualized representations; and (3) a \textbf{Hash Decoder} that auto-regressively reconstructs the multi-ID signature of the next token.

\subsection{Hash Encoder}

\paragraph{Multi-Hash Indexing.}
To map each discrete token $w$ into the hash signature space, we use $H$ independent hash functions: $\mathcal{H}_1(w), \mathcal{H}_2(w), \dots, \mathcal{H}_H(w)$.
For any non-padding token, the $i$-th coordinate of the signature is computed using the non-cryptographic MurmurHash3 (MMH3) algorithm \cite{senuma2025mmh3}. The algorithm uses iterative bitwise multiplication and rotation to diffuse the input, ensuring a single-bit change yields a uniform, randomized hash. We use a hash function-specific $\mathrm{seed}_i$:
\begin{align*}
\mathcal{H}_i(w) &= \mathrm{MMH3}(w, \mathrm{seed}_i)\\
&\pmod{B-1} + 1 \quad \text{if } w \neq \mathrm{pad}.
\end{align*}
\noindent $B$ denotes the number of discrete hash buckets per function.
$\mathcal{H}_i(\mathrm{pad}) = 0$ is reserved across $\mathcal{H}$ to denote the padding token. To guarantee that every token in the vocabulary maps to a collision-free multi-hash ID, we use an iterative rehashing strategy. If a new token generates a signature that conflicts with an existing vocabulary entry, we incrementally modify the seed of the final hash function ($\mathrm{seed}_H$) until an unused signature is found.

\paragraph{Gated Compositional Embedding.}
We allocate $H$ separate embedding matrices $\mathbf{E}^{(i)} \in \mathbb{R}^{B \times d}$ (where $i \in \{1, \dots, H\}$) to each hash coordinate $\mathcal{H}$, where $d$ represents the backbone hidden dimension.
Because hash buckets are shared globally per function, overlapping tokens that are semantically unrelated inevitably collide. Inspired by the multi-embedding approach of \citet{guo2023embedding}, we resolve this ambiguity by compressing the coordinates into a unified token representation via a context-aware compositional gate.
To determine the contribution of each bucket, the individual hash embeddings pass through a feed-forward bottleneck network with a compression dimension $d_z \ll d$, followed by softmax normalization. Finally, a linear adapter matrix $\mathbf{W}_s$ projects the combined representation into the latent space of the sequence processing backbone:
\begin{gather*}
\mathbf{e} = \left( \sum_{i=1}^H \tilde{\alpha}_i \mathbf{E}^{(i)}_{[\mathcal{H}_i(w), :]} \right) \mathbf{W}_{s}, \\
\tilde{\alpha}_i = \frac{\exp^{\alpha_i}}{\sum_{j=1}^H \exp^{\alpha_j}}, ~\alpha_i = \sigma \left( \mathbf{E}^{(i)}_{[\mathcal{H}_i(w), :]} \mathbf{W}_{1} \right) \mathbf{W}_{2}.
\end{gather*}
\noindent where $\mathbf{W}_s \in \mathbb{R}^{d \times d}$ represents the structural adapter, $\mathbf{W}_1 \in \mathbb{R}^{d \times d_z}$ and $\mathbf{W}_2 \in \mathbb{R}^{d_z \times 1}$ sequentially project the representations to an activation scaler with an non-linear activation above the intermediate bottleneck mainifold. Given an input sequence of tokens $(w_1, w_2, \dots, w_n)$, the Hash Encoder processes each token $w_i$ to its corresponding hash embedding $\mathbf{e}_i \in \mathbb{R}^d$.

\subsection{Sequence Processing Backbone}

$\mathbf{X} = [ \mathbf{e}_1, \mathbf{e}_2, \dots, \mathbf{e}_n]^\top$ where $\mathbf{X} \in \mathbb{R}^{n \times d}$ passes through a standard stack of $L$  Transformer layers. 
At each time-step $t$, the final layer of the backbone emits a contextualized latent vector $\mathbf{h}_t \in \mathbb{R}^d$: $\mathbf{h}_t = \mathrm{Transformer}(\mathbf{X}_{[:t,:]})$.
Subsequently, $\mathbf{h}_t$ is fed to the Hash Decoder.

\subsection{Hash Decoder}
To map the continuous hidden states back into the multi-hash ID signature space, we implement an auto-regressive \textit{Cascaded Predictor}. This module iteratively refines its index predictions.
It functions as a structured error-correcting system where preceding hashing choices directly constrain and contextualize subsequent token-signature inferences. To anchor this iterative cascade to the sequence context for predicting the next token $w_{t+1}$, the decoder initializes its prediction loop with the final hidden state of the backbone at the current time-step $t$. Thus, the root state $\mathbf{c}^{(1)} \in \mathbb{R}^d$ for the logit computation cascade is defined as $\mathbf{c}^{(1)} = \mathbf{h}_t$.

\paragraph{Logit Computation.} 
For each sequential hash head $i \in \{1, \dots, H\}$, the decoder projects the current latent hash state $\mathbf{c}^{(i)} \in \mathbb{R}^{d}$ to compute a logit distribution $\mathbf{o}^{(i)} \in \mathbb{R}^{B}$ over the physical bucket allocations. This is executed by slicing a dedicated subspace of the hash head weights $\mathbf{W}_{o}^{(i)} \in \mathbb{R}^{d \times B}$:
\begin{equation*}
\mathbf{o}^{(i)} =  \mathbf{W}_{o}^{(i)\top}\mathbf{c}^{(i)},
\end{equation*}
where $\mathbf{W}_{o}^{(i)}$ defines the localized parameter weight matrix assigned exclusively to the $i$-th coordinate bucket array. Following \citet{press-wolf-2017-using}, we tie the input and output embedding weights by setting $\mathbf{W}_{o}^{(i)\top}=\mathbf{E}^{(i)}$.

\paragraph{Soft Embedding Retrieval.} 
For all non-terminal heads ($i < H$), the decoder extracts a continuous representation of its prediction to pass down the cascade. We first compute a normalized probability distribution $\mathbf{p}^{(i)} \in \mathbb{R}^{B}$ across the bucket candidates via a softmax operation:
\begin{equation*}
\mathbf{p}^{(i)} = \mathrm{Softmax}(\mathbf{o}^{(i)}).
\end{equation*}
Using a hard discrete index during intermediate steps disrupts differentiability. Instead, the module computes a soft bucket embedding $\mathbf{e}^{(i)} \in \mathbb{R}^{d}$ as the expected value of the shared embedding matrix $\mathbf{E}^{(i)}$, weighted by the logit probabilities:
\begin{equation*}
\mathbf{e}^{(i)} = \mathbf{E}^{(i)\top}\mathbf{p}^{(i)}
\end{equation*}

\paragraph{State Update via Cascade Mixer.} 
To propagate the contextualized trajectory to the next signature head, the internal hash state requires an update. We achieve this using a recursive cascade mixer inspired by tree-structured recursive networks \cite{socher2011parsing, goller1996learning}. This block concatenates the existing state with the retrieved soft embedding and routes the joint representation through a bottleneck layer. Augmented with a structural residual connection, the subsequent state formulation $\mathbf{c}^{(i+1)}$ is defined as:\looseness=-1
\begin{equation*}
\mathbf{c}^{(i+1)} = \mathbf{c}^{(i)} +\mathbf{W}_{up}^{(i)\top}\sigma(\mathbf{W}_{dn}^{(i)\top}\begin{bmatrix} \mathbf{c}^{(i)} \\ \mathbf{e}^{(i)}\end{bmatrix}),
\end{equation*}
\noindent where $\sigma$ is an element-wise activation, and $\mathbf{W}_{dn}^{(i)} \in \mathbb{R}^{2d \times b}$ and $\mathbf{W}_{up}^{(i)} \in \mathbb{R}^{b \times h}$ represent the low-rank projection weights of the bottleneck layer.

\subsection{Probability Modeling}

Standard autoregressive LMs decompose the joint probability of the input  into a product of conditional probabilities.
In contrast, \textsc{MultiHashFormer} leverages the outputs of the Hash Decoder to factorize this token-level prediction.
Specifically, the normalized distributions ($\mathbf{p}^{(1)}, \dots, \mathbf{p}^{(H)}$) produced iteratively by the cascaded predictor serve directly as the hash-specific conditional probabilities $P(\mathcal{H}_i(w_t) \mid \cdot)$ for each signature coordinate $i \in \{1, \dots, H\}$.

\paragraph{Training Phase.}
The Hash Decoder maps tokens into a virtual vocabulary space $\mathcal{V}_{virt}$ defined by the total combinations of physical bucket allocations, where $|\mathcal{V}_{virt}| = B^H$. Because the actual language vocabulary is highly compact relative to this space, it forms a strict subset: $\mathcal{V}_{actl} \subset \mathcal{V}_{virt}$ and $|\mathcal{V}_{virt}| \gg |\mathcal{V}_{actl}|$. 
Consequently, we allow predictions over coordinate combinations that might not map to actual entries in the language vocabulary. This unconstrained formulation simplifies the optimization objective.
The training probability distribution for a given token $w$ is thus optimized via the product of its independent coordinate probabilities:\looseness=-1
\begin{equation*}
P_{train}(w \mid \cdot) = \prod_{i=1}^{H} P(\mathcal{H}_i(w) \mid \cdot).
\end{equation*}

\paragraph{Inference Phase.}
Allowing unconstrained decoding at inference could cause the model to generate coordinate combinations that do not map to actual semantic entries, leaking probability mass into invalid signatures. To guarantee the generation of valid multi-hash ID sequences, we explicitly exclude all unassigned signature during inference. This requires re-normalizing the probability distribution strictly over the true token vocabulary $\mathcal{V}_{actl}$:
\begin{align*}
P_{inf}(w \mid \cdot) = \frac{\prod_{i=1}^{H} P(\mathcal{H}_i(w) \mid \cdot)}{\sum_{w' \in \mathcal{V}_{actl}} \prod_{i=1}^{H} P(\mathcal{H}_i(w') \mid \cdot)}
\end{align*}
\noindent By accumulating the log-probabilities of individual hash IDs, the final step is a standard softmax normalization over the valid token space $\mathcal{V}_{actl}$.

\section{Experimental Setup}

\subsection{Models} 
We evaluate \textsc{MultiHashFormer} models against standard autoregressive LMs across 100M, 1B and 3B parameter scales. All configurations use a decoder-only Transformer backbone based on the Qwen3~\cite{yang2025qwen3technicalreport} architecture.
We use Mistral-7B-v0.3 \cite{jiang2023mistral7b}, a predominantly English BPE tokenizer with a 32K vocabulary.

\paragraph{Baselines.}
We use two baseline configurations:
(1) A causal LM with a conventional embedding matrix $\mathbf{E} \in \mathbb{R}^{|\mathcal{V}| \times d}$ and an equivalent LM head $\mathbf{W}_{o} \in \mathbb{R}^{d \times |\mathcal{V}|}$ (\textbf{Standard}); (2) a variant of Standard with $k$ additional transformer layers to match the total parameter count of \textsc{MultiHashFormer} (\textbf{Standard+$k$L}).
This allows us to verify that any performance improvements in \textsc{MultiHashFormer} do not result from increased parameter capacity.\footnote{Due to computational resource constraints, we consider this variant for 100M and 1B scales only.} Embedding and LM head weights are tied similar to the \textsc{MultiHashFormer}. We match the hidden dimension $d$ and the number of attention heads to the \textsc{MultiHashFormer}.

\begin{table*}[!t]
\centering
\small
\resizebox{\linewidth}{!}{
\begin{NiceTabular}{clrrcccccccccc}
\toprule
& & \multicolumn{2}{c}{\textbf{\#Param}} &\textbf{Lang. Model.} & \multicolumn{5}{c}{\textbf{Commonsense Reasoning}} & \multicolumn{3}{c}{\textbf{Reading Comprehension}} & \textbf{CR+RC}\\
\cmidrule(lr){5-5}   
\cmidrule(lr){6-10}   
\cmidrule(lr){11-13}  
\cmidrule(lr){14-14} 

& \textbf{Model} & \textbf{Emb} & \textbf{Dec} &\textbf{LAMBADA} & \textbf{ARC-E} & \textbf{COPA} & \textbf{OBQA} & \textbf{PIQA} & \textbf{HellaSwag} & \textbf{RACE} & \textbf{SciQ} & \textbf{SIQA} & \textbf{ReCoRD}\\
\midrule
& Rnd. Guess & - & - & - & 25.00 & 50.00 & 25.00 & 50.00 & 25.00 & 25.00 & 25.00 & 25.00 & 19.10\\
\midrule

\multirow{4}{*}{\rotatebox[origin=c]{90}{100M}} & Standard & 25M & 87M & 15.33 & 41.33 & \textbf{64.00} & 27.00 & 58.87 & 28.59 & 26.41 & 59.00 & 35.98 & 49.47\\
& Standard+4L & 25M & 116M & \textbf{19.64} & 45.08 & \textbf{64.00} & 27.00 & 60.17 & \textbf{29.99} & 26.79 & \textbf{62.60} & 36.80 & \textbf{53.08}\\
\noalign{\vskip\aboverulesep}\cdashline{2-15}[2pt/1.2pt]\noalign{\vskip\belowrulesep}

& MHF (H3B10K) & 25M & 87M & 16.40 & 44.91 & 61.00 & 26.40 & \textbf{60.23} & 27.81 & 27.18 & 55.50 & \textbf{37.00} & 48.21\\

& MHF (H4B16K) & 52M & 87M & 18.79 & \textbf{45.24} & 62.00 & \textbf{27.60} & 59.03 & 27.99 & \textbf{28.42} & 58.00 & 36.03 & 50.11\\

\midrule

\multirow{4}{*}{\rotatebox[origin=c]{90}{1B}} & Standard & 65M & 0.9B & 30.41 & 59.85 & 64.00 & 31.60 & 65.78 & 38.99 & 29.38 & 69.90 & 39.10 & 63.20\\
& Standard+2L & 65M & 1.0B & 30.66 & 60.23 & 65.00 & 31.20 & 65.02 & 39.85 & 29.28 & \textbf{72.10} & 38.89 & 63.78\\
\noalign{\vskip\aboverulesep}\cdashline{2-15}[2pt/1.2pt]\noalign{\vskip\belowrulesep}

& MHF (H3B10K) & 65M & 0.9B & \textbf{35.78} & 61.95 & 65.00 & \textbf{34.00} & \textbf{68.17} & 40.45 & 29.00 & 68.30 & 40.48 & 64.51\\

& MHF (H4B16K) & 138M & 0.9B & 35.34 & \textbf{64.35} & \textbf{71.00} & 29.80 & 66.49 & \textbf{41.32} & \textbf{31.39} & 70.20 & \textbf{41.66} & \textbf{64.90}\\

\midrule

\multirow{2}{*}{\rotatebox[origin=c]{90}{3B}} & Standard & 65M & 2.8B & 28.64 & 60.23 & 65.00 & 31.80 & 65.83 & 39.65 & 29.76 & 64.80 & 38.28 & 62.26\\
\noalign{\vskip\aboverulesep}\cdashline{2-15}[2pt/1.2pt]\noalign{\vskip\belowrulesep}
& MHF (H4B16K) & 138M & 2.8B & \textbf{37.26} & \textbf{66.88} & \textbf{68.00} & \textbf{34.60} & \textbf{68.39} & \textbf{42.89} & \textbf{30.91} & \textbf{70.60} & \textbf{40.74} & \textbf{64.13}\\

\bottomrule

\end{NiceTabular}
}
\caption{Performance of standard and our \textsc{MultiHashFormer} models. Bold denotes best performance.}
\label{table:performance_1B_3B}

\end{table*}

\paragraph{MultiHashFormer.}
We use the same core sequence processing backbone to the baseline configurations. 
We denote variants as H$H$B$B$, where $H$ is the number of independent hash functions and $B$ is the physical bucket allocation per head.
We evaluate two configurations: \textbf{H3B10K} ($H=3$, $B=10,240$)\footnotemark[2], which directly matches the baseline parameter count, and \textbf{H4B16K} ($H=4$, $B=16,384$), our optimal configuration based on \S\ref{sec:analysis}.
For brevity, we fix the embedding matrices and projection layer sizes across all vocabulary settings.
The bottleneck projection dimension for both the Gated Compositional Embedding and the Cascade Mixer is set to $d_z=64$ (where $d_z \ll d$).

\subsection{Pre-training Data and Hyperparameters}
All models are pre-trained from scratch on a subset of English FineWeb-Edu~\citep{penedo2024fineweb}. Following Chinchilla scaling laws \citep{hoffmann2022an}, we adjust data volume based on model scale. The 100M models were trained on 10B tokens, and the 1B and 3B models on 100B tokens. We use a global batch size of 256 with a 2,048-token context window, following \citet{yamaguchi-etal-2026-enhancing} and \citet{xue-etal-2025-deconstructing}. See Table~\ref{table:hyperparams_pt} in Appendix~\ref{appendix:hyperparam} for details.\looseness=-1

\subsection{Evaluation Benchmarks}\label{section:evaluation_benchmark}

\paragraph{Core capabilities.} 
We include LAMBADA~\citep{paperno-etal-2016-lambada} for Language Modeling; Commonsense Reasoning (CR) using ARC-Easy~\citep{bhakthavatsalam2021think}, COPA~\citep{gordon-etal-2012-semeval}, OBQA~\citep{mihaylov-etal-2018-suit}, PIQA~\citep{bisk2020piqa}, HellaSwag~\citep{zellers-etal-2019-hellaswag}; and Reading Comprehension (RC) with RACE~\citep{lai-etal-2017-race}, SciQ~\citep{welbl-etal-2017-crowdsourcing}, SIQA~\citep{sap-etal-2019-social}. We evaluate CR + RC using ReCoRD~\citep{zhang2018record}.\footnote{We report normalized accuracy for ARC-E, HellaSwag, SIQA, OBQA, and SciQ; F1 scores for ReCoRD; and standard accuracy for all remaining tasks. We use a five-shot setting for ARC-E, SIQA, and OBQA, while the rest are evaluated in a zero-shot setting.}

\paragraph{Rare Words.} We hypothesize that forcing sparse tokens to share representational capacity would alleviate the cold start problem of rare vocabulary words and further enable the model to learn their semantic representations more effectively. To investigate if the shared bucket mechanism of \textsc{MultiHashFormer} improve rare word representations, we evaluate 1B models on the Card-660 dataset \citep{pilehvar-etal-2018-card}. This dataset contains word pairs (rare-rare or rare-frequent) with human-annotated similarity scores. We compute the cosine similarity between the words using the final and second-to-last hidden states corresponding to the last token. Finally, we measure the correlation between these model-derived scores and the human annotations.

\subsection{Vocabulary Expansion}
As mentioned in \S\ref{sec:introduction}, a key theoretical advantage of \textsc{MultiHashFormer} is its ability to expand its vocabulary capacity without increasing its parameter footprint.
To evaluate this capability, we test the models on their ability to integrate new tokens during multilingual vocabulary expansion.

\paragraph{Continual Pre-training Data.}
We continue pre-training (CPT) models on a multilingual corpus containing 6B tokens evenly split across Arabic, Chinese, and Hindi from FineWeb2~\citep{penedo2025fineweb2pipelinescale}.
We also include an extra 2B English tokens from FineWeb-Edu~\citep{penedo2024fineweb} to mitigate catastrophic forgetting.
Following \citet{yao-etal-2021-adapt}, we add 5K tokens per language, yielding a final vocabulary of 48K tokens. 

\paragraph{Model Configuration.}
For \textsc{MultiHashFormer}, we expand the vocabulary size by registering new unique virtual (multi-hash ID) signatures \textit{without adding new parameters}. For baselines, the weights of the new tokens within the embedding matrix are initialized using mean initialization, a standard protocol where each new token receives the average embedding of its corresponding source tokens from the original tokenizer~\citep{mundra-etal-2024-empirical,tejaswi-etal-2024-exploring,yamaguchi2025adapting,10.1162/COLI.a.581}.
We tune embeddings, LM heads and the first two and last two Transformer layers, following~\citet{remy2024transtokenization,owodunni2025continually,nakash2025adaptivocab}. See Table~\ref{table:hyperparams_cpt} (Appendix~\ref{appendix:hyperparam}) for full details.

\paragraph{Evaluation.}
We use five multilingual tasks from MuBench~\citep{han-etal-2026-mubench}, consisting of HellaSwag, TruthfulQA~\citep{lin-etal-2022-truthfulqa}, StoryCloze~\citep{mostafazadeh-etal-2016-corpus}, and MMLU~\citep{hendrycks2020measuring}.\footnote{We only include tasks where all Standard baselines outperform random guessing.
For MMLU, we apply this filtering criterion at the subtask level and report the average over the retained subtasks.
We report zero-shot accuracy for MMLU and normalized accuracy for the remaining tasks.}
We also include the English benchmarks from~\S\ref{section:evaluation_benchmark} to track catastrophic forgetting.

\section{Results}

\subsection{Core Capabilities} \label{subsec:core_results}
Table~\ref{table:performance_1B_3B} presents the performance of the baselines alongside \textsc{MultiHashFormer} (MHF) models across the 100M, 1B, and 3B parameter scales.

\paragraph{Efficacy while scaling parameter counts.} We first observe that as the model capacity increases from 1B to 3B parameters, MHF (H4B16K) consistently outperforms the baseline on 9 out of 10 tasks. Specifically, on the LAMBADA task, MHF (H4B16K) offers substantial gains of 4.93\% and 8.62\% over the Standard baseline (30.41\% and 28.64\%) at the 1B and 3B scales, respectively. Although the standard LM baseline achieves higher scores on OBQA at the 1B scale, the overall results indicate the efficacy of \textsc{MultiHashFormer} in capturing contextual dependencies.

\paragraph{Gains under strict parameter matching.} When matching the parameter count at the 1B scale, \textsc{MultiHashFormer} variants offer superior performance compared to  baselines.
Specifically, MHF (H3B10K) and MHF (H4B16K) outperform Standard and Standard+2L baselines on 8 out of 10 tasks, respectively.
Notably, MHF (H4B16K) achieves 64.90 on ReCoRD, surpassing the 63.78 of Standard+2L. Similarly, MHF (H3B10K) outperforms Standard on COPA with 71.00 vs. 65.00.
These results suggest that decoupling the vocabulary is effective. Allocating more parameters to the embeddings improves performance more than adding more layers to the model on larger scales.\looseness=-1

\begin{figure*}[!t]
    \centering
    \includegraphics[width=1.0\linewidth]{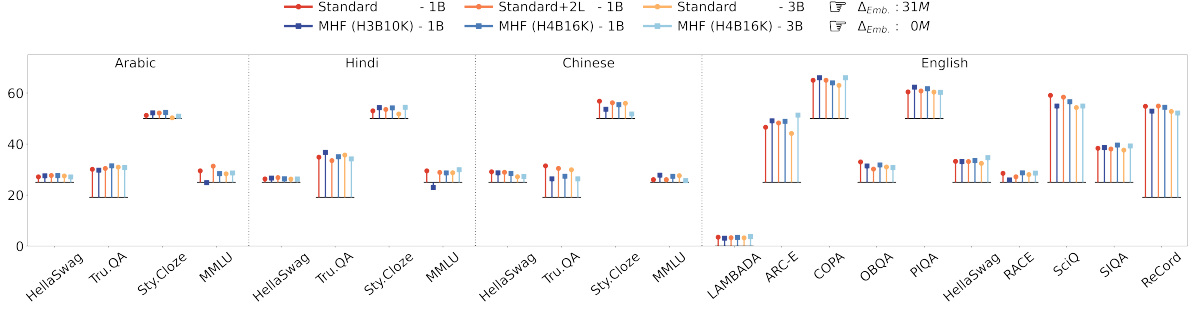}
    \caption{\textsc{MultiHashFormer} (blue shades, diamond marker) are similar or better than Standard (orange shades, circle marker) at 1B and 3B scales after expanding the vocabulary from 32K to 48K without adding a single new parameter. Black horizontal lines denote random baseline performance per task.}
    \label{fig:ve}
\end{figure*}

\paragraph{Per-task performance.} Looking at specific tasks, we note a performance trade-off based on the hash configuration.
On OBQA, models with fewer hash buckets achieve higher accuracy.
This reversal likely stems from the long-tail distribution of the dataset. A large bucket allocation fragments the embedding space for sparse vocabulary items, which degrades the representation of rare factual tokens.
Conversely, reading comprehension requires high representational capacity to prevent coordinate-level collisions between tokens across complex contexts.
At the 1B scale, the lower-capacity MHF (H3B10K) variant underperforms the Standard baseline on SciQ (68.30\% vs. 69.90\%) and RACE (29.00\% vs. 29.38\%), although it outperforms the baseline on SIQA (40.48\% vs. 39.10\%).
Expanding the physical bucket allocation to the MHF (H4B16K) configuration improves these scores to 70.20\% on SciQ and 31.39\% on RACE, surpassing the Standard baseline.

\textsc{MultiHashFormer}'s strong performance on LM and reasoning tasks stems primarily from its ability to mitigate the softmax bottleneck~\citep{yang2017breaking}. Hash decoder effectively lifts the rank upper-bound of the log-probability matrix estimation in language modeling to $\mathrm{min}(B,H\times{d})$, which is substantially higher then $d$ for the standard LM head. Its mathematical derivation is further detailed in Appendix~\ref{appendix:softmax_bottleneck}. This advantage is especially evident in LAMBADA, which requires predicting specific, context-dependent words, and HellaSwag, which tests narrative logic. Our model effectively distinguishes between high-probability candidates that Standard baselines often conflate.

\paragraph{Performance saturation at 100M.} At the 100M scale, the decoder size is more important than the vocabulary representation. \textsc{MultiHashFormer} variants do not consistently outperform Standard baselines, especially on COPA, HellaSwag, and ReCoRD. Instead, increasing decoder depth by 4 layers is better across 7 tasks. We attribute this to three constraints of smaller and shallower architectures: (1) restricted hidden dimensions conflate distinct tokens with similar hash signatures; (2) shallower depth prevents the model from contextualizing and disambiguating these embeddings; and (3) narrower hidden states restrict the rank upper bound (see our theoretical analysis in Appendix~\ref{appendix:softmax_bottleneck}), degrading log-probability estimation. Consequently, \textsc{MultiHashFormer} works more effectively with larger decoders.

\begin{table}[!t]
\centering
\small
\resizebox{1.0\linewidth}{!}{
\begin{NiceTabular}{lcccc}
\toprule
\textbf{Model} & \multicolumn{2}{c}{\textbf{Last Hidden States}} & \multicolumn{2}{c}{\textbf{Second-to-Last Hidden States}}\\
& \textbf{Pearson $r$ $\uparrow$} & \textbf{Spearman $\rho$ $\uparrow$} & \textbf{Pearson $r$ $\uparrow$} & \textbf{Spearman $\rho$ $\uparrow$}\\
\midrule

Starndard & 0.20 & 0.22 & 0.26 & 0.25\\
MHF (H3B10K) & \textbf{0.23} & \textbf{0.23} & \textbf{0.32} & \textbf{0.29}\\
\noalign{\vskip\aboverulesep}\cdashline{1-5}[2pt/1.2pt]\noalign{\vskip\belowrulesep}

Standard+2L & \textbf{0.22} & 0.22 & 0.29 & 0.29\\
MHF (H4B16K) & 0.20 & \textbf{0.26} & \textbf{0.30} & \textbf{0.32}\\

\bottomrule

\end{NiceTabular}
}
\caption{Correlations between human annotations and cosine similarities computed from the last two hidden states of the 1B models on the Card-660 dataset. 
}
\label{table:card660}

\end{table}

\subsection{Rare Word Representations}

Table~\ref{table:card660} shows the correlation coefficients on the Card-660 dataset. The \textsc{MultiHashFormer} variants consistently outperform Standard models while maintaining an equivalent parameter count. We find that \textsc{MultiHashFormer} is better at identifying semantically equivalent word pairs (see examples in Appendix~\ref{appendix:card660_appx}). This performance gap is particularly evident when analyzing hidden states from the second-to-last decoder layer, where representations are less biased toward the specific training tasks. 
Our model effectively captures semantic representations of rare vocabulary items by forcing sparse tokens to share representational capacity.

\subsection{Vocabulary Expansion}

Figure~\ref{fig:ve} illustrates the performance of the adapted baselines compared to the adapted \textsc{MultiHashFormer} models at 1B and 3B parameter scales across Arabic, Hindi, Chinese, and English. Notably, both the 1B and 3B MHF (H4B16K) models consistently outperform Standard on 12 and 13 out of 22 multilingual tasks, respectively. They achieve this without the additional 31-million-parameters needed to accommodate 15K new tokens in the expanded vocabulary of the Standard baselines. 
We further find that MHF (H3B10K) and MHF (H4B16K) perform comparably (difference smaller than 1\%) on 15 out of 22 tasks. This indicates that performance remains robust during vocabulary expansion across different hash configurations. Furthermore, under a strict parameter-count matching constraint prior to vocabulary expansion, MHF (H3B10K) and MHF (H4B16K) achieve better or comparable performance to the Standard models on the majority of English tasks (6 and 9 out of 10, respectively). This suggests that \textsc{MultiHashFormer} does not suffer from  more catastrophic forgetting relative to  Standard. It successfully supports a 46.9\% larger multilingual vocabulary \textit{without a single additional parameter} while preserving its core capabilities in English.

\section{Analysis} \label{sec:analysis}

We further analyze the core architectural components of \textsc{MultiHashFormer}. Due to computational resource constraints, we conduct this analysis at the 1B parameter trained up to 20B tokens.

\paragraph{Single vs. Multi-hash ID.}
To demonstrate the necessity of Multi-ID signatures, we evaluate three different \textsc{MultiHashFormer} configurations (H4B4K, H4B8K, and H4B16K) against their Single-ID counterparts (H1B4K, H1B8K, and H1B16K). Table~\ref{table:single_multi} presents the results on LAMBADA. Models with Multi-ID signatures consistently outperform those using Single-ID signatures across all configurations (B4K, B8K, and B16K). 
Specifically, the performance difference is higher at lower bucket capacities (B4K), where H4B4K achieves 30.27\% accuracy compared to 4.29\% for H1B4K. Expanding the capacity to B16K with Single-ID only improves accuracy to 14.30\%. 
These results suggest that a signature collision severely degrades the core capability of the model. Employing Multi-ID signatures to prevent such collisions is more effective than increasing the number of hash buckets.

\begin{table}[!t]
\centering
{\tiny
\resizebox{0.7\linewidth}{!}{
\begin{NiceTabular}{lrrr}
\toprule
MHF & \textbf{B4K} & \textbf{B8K} & \textbf{B16K}\\
\midrule
\textbf{H1} & 4.29	& 8.77 & 14.30\\
\textbf{H4} & 30.27	& 30.47 & 30.91\\
\bottomrule
\end{NiceTabular}%
}
}
\caption{LAMBADA accuracy across different MHF (H4B$B$) configurations against their corresponding variants H1B$B$ using Single-ID signature.\tablefootnote{H1B$B$ use single value for 4 coordinates to strictly match the number of parameters in H4B$B$ for a fair comparison.}}
\label{table:single_multi}
\end{table}

\begin{figure}[!t]
\centering
\includegraphics[width=1.0\linewidth]{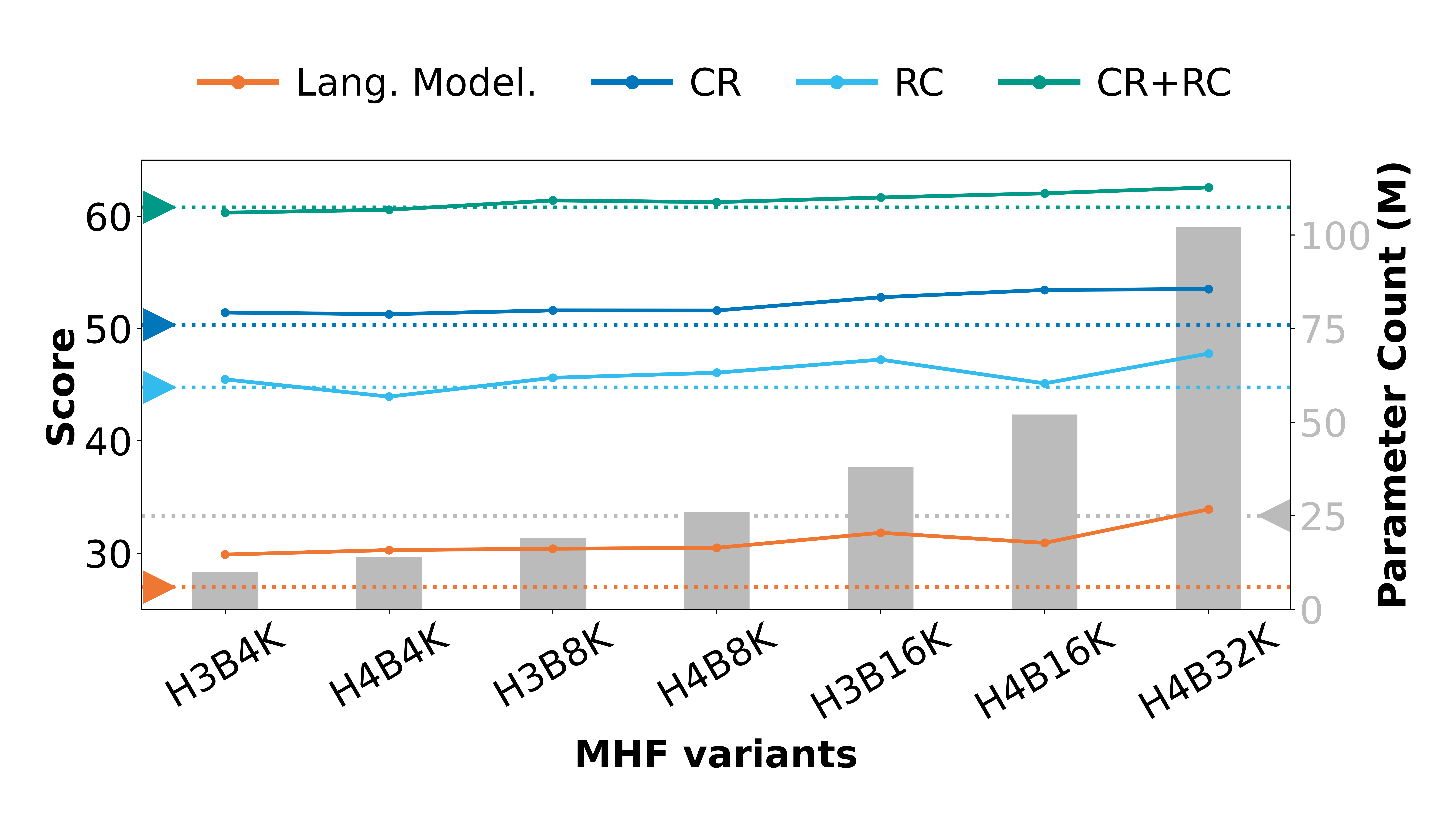}
\caption{Average scores on four core capabilities across 1B \textsc{MultiHashFormer} variants at  against their parameter counts in embeddings (vertical bars). Horizontal dotted lines represent the Standard baseline.}\label{fig:random_ablation}
\end{figure}

\paragraph{Varying Hash Functions and Bucket Size.} 
To evaluate the impact of hash configuration on parameter efficiency and downstream performance, we test various combinations of hash functions ($H$) and bucket sizes ($B$). Figure~\ref{fig:random_ablation} illustrates average scores on four core capabilities of \textsc{MultiHashFormer} variants against their embedding parameter counts, across different combinations of $H$ and $B$.
All \textsc{MultiHashFormer} configurations consistently outperform Standard on Language Modeling and Commonsense Reasoning, including variants that utilize strictly smaller embedding matrices, such as H3B4K, H4B4K, and H3B8K.
Although scaling both $H$ and $B$ sharply increases the embedding parameter count, the corresponding accuracy scores remain stable.
Specifically, the best-performing variant improves over the lowest by only 4\%, despite a tenfold increase in parameter count from 10M to 102M. Consequently, the H4B16K variant achieves an optimal balance between parameter efficiency and predictive accuracy.
We therefore adopt H4B16K as the primary configuration for the 1B and 3B scaling experiments presented in \S\ref{subsec:core_results}.

\begin{table}[!t]
\centering
\small
\resizebox{1.0\linewidth}{!}{
\begin{NiceTabular}{lccccc}
\toprule
\textbf{LSH : MMH3} & \textbf{0 : 4} & \textbf{1 : 3} & \textbf{2 : 2} & \textbf{3 : 1} & \textbf{4 : 0}\\
\midrule
\textbf{LAMBADA} & 30.91 & 30.60 & 30.80 & 31.40 & 31.36\\
\bottomrule
\end{NiceTabular}%
}
\caption{LAMBADA accuracy across different MHF(H4B16K) variants while replacing a different number of MMH3 functions to LSH.}
\label{table:random_lsh}
\end{table}

\paragraph{MMH3 vs. LSH.}
We also investigate whether incorporating Locality-Sensitive Hashing \citep[LSH]{datar2004locality} enhances the general capability of the model by capturing morphological similarity.
Because LSH maps structurally similar input features to identical hash buckets with high probability, applying it to character-level trigrams tends to force subwords with similar spelling to share embedding representations.
To evaluate this mechanism, we pre-train three additional MHF (H4B16K) variants from scratch at a 1B-parameter scale, keeping the dataset and hyperparameters identical.
Table~\ref{table:random_lsh} presents the performance of these models on LAMBADA.
Increasing the proportion of LSH functions while fixing the total number of hash functions does not yield a consistent accuracy improvement.
This outcome suggests that explicitly injecting a morphological inductive bias is not necessary given the inherent computational complexity of LSH compared to MMH3.
The standard deterministic random hashing of MMH3 provides sufficient representational flexibility for the network to learn subword relationships entirely end-to-end.

\section{Conclusion}

We introduced \textsc{MultiHashFormer}, a generative framework that leverages multi-hash structures to bypass traditional vocabulary bottlenecks.  Empirical evaluation demonstrates that it offers improvements both in core capabilities and rare word representations. Furthermore, the framework enables seamless vocabulary expansion without requiring additional parameters. In future work, we plan to extend \textsc{MultiHashFormer} to cross token boundary multi-hash generation, to improve multi-token prediction~\citep{gloeckle2024better}.

\section*{Limitations}

\paragraph{Model size.} Due to computational resource constraints, we restrict our evaluation to models at the 100M, 1B and 3B scales. Following the protocol of~\citet{cheng-etal-2024-instruction}, 1B and 3B models were trained from scratch within a standard academic budget of 100B tokens. While \textsc{MultiHashFromer} shows an upward performance trajectory with model size, exploring larger scales (e.g., 7B+ parameters) remains a valuable direction for institutions with access to the required computing resources.

\paragraph{Single seed.} Due to the high computational cost of pre-training 1B and 3B parameter models from scratch for 100B tokens, conducting multiple training runs was prohibitive under our academic infrastructure constraints. Consequently, our empirical results are based on a single seed per configuration. However, the consistent performance gains observed across both the 1B and 3B scales demonstrate the robustness of the \textsc{MultiHashFormer} framework. This consistency indicates that the improvements introduced by our Hash Encoder and Hash Decoder stem from inductive bias rather than random sampling variance. Furthermore, adopting the established hyperparameter protocols of~\citet{li2025predictable} and~\citet{cheng-etal-2024-instruction}, we ensure standard and reproducible training dynamics.

\section*{Acknowledgments}
We would like to thank Maggie Mi for the valuable feedback.
We acknowledge (1) IT Services at the University of Sheffield for the provision of services for high-performance computing; (2) the use of the University of Oxford Advanced Research Computing (ARC) facility; (3) the Isambard-AI National AI Research Resource (AIRR), which is operated by the University of Bristol and is funded by the UK Government’s Department for Science, Innovation and Technology (DSIT) via UK Research and Innovation; and the Science and Technology Facilities Council [ST/AIRR/I-A-I/1023]; and (4) the EuroHPC Joint Undertaking for awarding us access to Leonardo, hosted by CINECA (Italy).
AY is supported by the Engineering and Physical Sciences Research Council (EPSRC) [grant number EP/W524360/1] and the Japan Student Services Organization (JASSO) Student Exchange Support Program (Graduate Scholarship for Degree Seeking Students).

\bibliography{custom,anthology-1,anthology-2}

\begin{thebibliography}{63}
\providecommand{\natexlab}[1]{#1}

\bibitem[{Abdin et~al.(2024)Abdin, Aneja, Behl, Bubeck, Eldan, Gunasekar, Harrison, Hewett, Javaheripi, Kauffmann, Lee, Lee, Li, Liu, Mendes, Nguyen, Price, de~Rosa, Saarikivi, Salim, Shah, Wang, Ward, Wu, Yu, Zhang, and Zhang}]{abdin2024phi4technicalreport}
Marah Abdin, Jyoti Aneja, Harkirat Behl, Sébastien Bubeck, Ronen Eldan, Suriya Gunasekar, Michael Harrison, Russell~J. Hewett, Mojan Javaheripi, Piero Kauffmann, James~R. Lee, Yin~Tat Lee, Yuanzhi Li, Weishung Liu, Caio C.~T. Mendes, Anh Nguyen, Eric Price, Gustavo de~Rosa, Olli Saarikivi, and 8 others. 2024.
\newblock \href {https://arxiv.org/abs/2412.08905} {Phi-4 technical report}.
\newblock \emph{Preprint}, arXiv:2412.08905.

\bibitem[{Bhakthavatsalam et~al.(2021)Bhakthavatsalam, Khashabi, Khot, Mishra, Richardson, Sabharwal, Schoenick, Tafjord, and Clark}]{bhakthavatsalam2021think}
Sumithra Bhakthavatsalam, Daniel Khashabi, Tushar Khot, Bhavana~Dalvi Mishra, Kyle Richardson, Ashish Sabharwal, Carissa Schoenick, Oyvind Tafjord, and Peter Clark. 2021.
\newblock \href {https://arxiv.org/abs/2102.03315} {Think you have solved direct-answer question answering? try {ARC-DA}, the direct-answer {AI2} reasoning challenge}.
\newblock \emph{Preprint}, arXiv:2102.03315.

\bibitem[{Bisk et~al.(2020)Bisk, Zellers, Le~Bras, Gao, and Choi}]{bisk2020piqa}
Yonatan Bisk, Rowan Zellers, Ronan Le~Bras, Jianfeng Gao, and Yejin Choi. 2020.
\newblock \href {https://doi.org/10.1609/aaai.v34i05.6239} {{PIQA}: Reasoning about physical commonsense in natural language}.
\newblock In \emph{Proceedings of the AAAI Conference on Artificial Intelligence}, volume~34, pages 7432--7439.

\bibitem[{Bricken and Pehlevan(2021)}]{bricken2021attention}
Trenton Bricken and Cengiz Pehlevan. 2021.
\newblock \href {https://proceedings.neurips.cc/paper_files/paper/2021/file/8171ac2c5544a5cb54ac0f38bf477af4-Paper.pdf} {Attention approximates sparse distributed memory}.
\newblock In \emph{Advances in Neural Information Processing Systems}, volume~34, pages 15301--15315. Curran Associates, Inc.

\bibitem[{Cheng et~al.(2024)Cheng, Gu, Huang, Bi, Huang, and Wei}]{cheng-etal-2024-instruction}
Daixuan Cheng, Yuxian Gu, Shaohan Huang, Junyu Bi, Minlie Huang, and Furu Wei. 2024.
\newblock \href {https://doi.org/10.18653/v1/2024.emnlp-main.148} {Instruction pre-training: Language models are supervised multitask learners}.
\newblock In \emph{Proceedings of the 2024 Conference on Empirical Methods in Natural Language Processing}, pages 2529--2550, Miami, Florida, USA. Association for Computational Linguistics.

\bibitem[{Clark et~al.(2022)Clark, Garrette, Turc, and Wieting}]{clark-etal-2022-canine}
Jonathan~H. Clark, Dan Garrette, Iulia Turc, and John Wieting. 2022.
\newblock \href {https://doi.org/10.1162/tacl_a_00448} {Canine: Pre-training an efficient tokenization-free encoder for language representation}.
\newblock \emph{Transactions of the Association for Computational Linguistics}, 10:73--91.

\bibitem[{Datar et~al.(2004)Datar, Immorlica, Indyk, and Mirrokni}]{datar2004locality}
Mayur Datar, Nicole Immorlica, Piotr Indyk, and Vahab~S. Mirrokni. 2004.
\newblock \href {https://doi.org/10.1145/997817.997857} {Locality-sensitive hashing scheme based on p-stable distributions}.
\newblock In \emph{Proceedings of the Twentieth Annual Symposium on Computational Geometry}, SCG '04, page 253–262, New York, NY, USA. Association for Computing Machinery.

\bibitem[{Deiseroth et~al.(2024)Deiseroth, Brack, Schramowski, Kersting, and Weinbach}]{deiseroth-etal-2024-free}
Bj{\"o}rn Deiseroth, Manuel Brack, Patrick Schramowski, Kristian Kersting, and Samuel Weinbach. 2024.
\newblock \href {https://doi.org/10.18653/v1/2024.emnlp-main.1217} {{T}-{FREE}: Subword tokenizer-free generative {LLM}s via sparse representations for memory-efficient embeddings}.
\newblock In \emph{Proceedings of the 2024 Conference on Empirical Methods in Natural Language Processing}, pages 21829--21851, Miami, Florida, USA. Association for Computational Linguistics.

\bibitem[{Ganchev and Dredze(2008)}]{ganchev-dredze-2008-small}
Kuzman Ganchev and Mark Dredze. 2008.
\newblock \href {https://aclanthology.org/W08-0804/} {Small statistical models by random feature mixing}.
\newblock In \emph{Proceedings of the {ACL}-08: {HLT} Workshop on Mobile Language Processing}, pages 19--20, Columbus, Ohio. Association for Computational Linguistics.

\bibitem[{Gloeckle et~al.(2024)Gloeckle, Youbi~Idrissi, Roziere, Lopez-Paz, and Synnaeve}]{gloeckle2024better}
Fabian Gloeckle, Badr Youbi~Idrissi, Baptiste Roziere, David Lopez-Paz, and Gabriel Synnaeve. 2024.
\newblock \href {https://proceedings.mlr.press/v235/gloeckle24a.html} {Better \& faster large language models via multi-token prediction}.
\newblock In \emph{Proceedings of the 41st International Conference on Machine Learning}, volume 235 of \emph{Proceedings of Machine Learning Research}, pages 15706--15734. PMLR.

\bibitem[{Goller and Kuchler(1996)}]{goller1996learning}
Christoph Goller and Andreas Kuchler. 1996.
\newblock \href {https://doi.org/10.1109/ICNN.1996.548916} {Learning task-dependent distributed representations by backpropagation through structure}.
\newblock In \emph{Proceedings of International Conference on Neural Networks (ICNN'96)}, volume~1, pages 347--352.

\bibitem[{Gordon et~al.(2012)Gordon, Kozareva, and Roemmele}]{gordon-etal-2012-semeval}
Andrew Gordon, Zornitsa Kozareva, and Melissa Roemmele. 2012.
\newblock \href {https://aclanthology.org/S12-1052/} {{S}em{E}val-2012 task 7: Choice of plausible alternatives: An evaluation of commonsense causal reasoning}.
\newblock In \emph{*{SEM} 2012: The First Joint Conference on Lexical and Computational Semantics {--} Volume 1: Proceedings of the main conference and the shared task, and Volume 2: Proceedings of the Sixth International Workshop on Semantic Evaluation ({S}em{E}val 2012)}, pages 394--398, Montr{\'e}al, Canada. Association for Computational Linguistics.

\bibitem[{Guo et~al.(2024)Guo, Pan, Wang, Chen, Jiang, and Long}]{guo2023embedding}
Xingzhuo Guo, Junwei Pan, Ximei Wang, Baixu Chen, Jie Jiang, and Mingsheng Long. 2024.
\newblock \href {https://proceedings.mlr.press/v235/guo24e.html} {On the embedding collapse when scaling up recommendation models}.
\newblock In \emph{Proceedings of the 41st International Conference on Machine Learning}, volume 235 of \emph{Proceedings of Machine Learning Research}, pages 16891--16909. PMLR.

\bibitem[{Han et~al.(2026)Han, Zhang, Chen, Binbinliu, Pechenizkiy, Fang, and Zheng}]{han-etal-2026-mubench}
Wenhan Han, Yifan Zhang, Zhixun Chen, Binbinliu, Mykola Pechenizkiy, Meng Fang, and Yin Zheng. 2026.
\newblock \href {https://doi.org/10.18653/v1/2026.findings-acl.794} {{M}u{B}ench: Assessment of multilingual capabilities of large language models across 61 languages}.
\newblock In \emph{Findings of the {A}ssociation for {C}omputational {L}inguistics: {ACL} 2026}, pages 16163--16192, San Diego, California, United States. Association for Computational Linguistics.

\bibitem[{Hendrycks et~al.(2021)Hendrycks, Burns, Basart, Zou, Mazeika, Song, and Steinhardt}]{hendrycks2020measuring}
Dan Hendrycks, Collin Burns, Steven Basart, Andy Zou, Mantas Mazeika, Dawn Song, and Jacob Steinhardt. 2021.
\newblock \href {https://openreview.net/forum?id=d7KBjmI3GmQ} {Measuring massive multitask language understanding}.
\newblock In \emph{International Conference on Learning Representations}.

\bibitem[{Hoffmann et~al.(2022)Hoffmann, Borgeaud, Mensch, Buchatskaya, Cai, Rutherford, de~Las~Casas, Hendricks, Welbl, Clark, Hennigan, Noland, Millican, van~den Driessche, Damoc, Guy, Osindero, Simonyan, Elsen, Vinyals, Rae, and Sifre}]{hoffmann2022an}
Jordan Hoffmann, Sebastian Borgeaud, Arthur Mensch, Elena Buchatskaya, Trevor Cai, Eliza Rutherford, Diego de~Las~Casas, Lisa~Anne Hendricks, Johannes Welbl, Aidan Clark, Thomas Hennigan, Eric Noland, Katherine Millican, George van~den Driessche, Bogdan Damoc, Aurelia Guy, Simon Osindero, Kar\'{e}n Simonyan, Erich Elsen, and 3 others. 2022.
\newblock \href {https://doi.org/10.52202/068431-2176} {An empirical analysis of compute-optimal large language model training}.
\newblock In \emph{Advances in Neural Information Processing Systems}, volume~35, pages 30016--30030. Curran Associates, Inc.

\bibitem[{Hu et~al.(2022)Hu, yelong shen, Wallis, Allen-Zhu, Li, Wang, Wang, and Chen}]{hu2021lora}
Edward~J Hu, yelong shen, Phillip Wallis, Zeyuan Allen-Zhu, Yuanzhi Li, Shean Wang, Lu~Wang, and Weizhu Chen. 2022.
\newblock \href {https://openreview.net/forum?id=nZeVKeeFYf9} {Lo{RA}: Low-rank adaptation of large language models}.
\newblock In \emph{International Conference on Learning Representations}.

\bibitem[{Hwang et~al.(2026)Hwang, Wang, and Gu}]{hwang2025dynamic}
Sukjun Hwang, Brandon Wang, and Albert Gu. 2026.
\newblock \href {https://openreview.net/forum?id=ZbfLR9NbNF} {Dynamic chunking for end-to-end hierarchical sequence modeling}.
\newblock In \emph{The Fourteenth International Conference on Learning Representations}.

\bibitem[{Jiang et~al.(2023)Jiang, Sablayrolles, Mensch, Bamford, Chaplot, de~las Casas, Bressand, Lengyel, Lample, Saulnier, Lavaud, Lachaux, Stock, Scao, Lavril, Wang, Lacroix, and Sayed}]{jiang2023mistral7b}
Albert~Q. Jiang, Alexandre Sablayrolles, Arthur Mensch, Chris Bamford, Devendra~Singh Chaplot, Diego de~las Casas, Florian Bressand, Gianna Lengyel, Guillaume Lample, Lucile Saulnier, Lélio~Renard Lavaud, Marie-Anne Lachaux, Pierre Stock, Teven~Le Scao, Thibaut Lavril, Thomas Wang, Timothée Lacroix, and William~El Sayed. 2023.
\newblock \href {https://arxiv.org/abs/2310.06825} {Mistral {7B}}.
\newblock \emph{Preprint}, arXiv:2310.06825.

\bibitem[{Kudo and Richardson(2018)}]{kudo-richardson-2018-sentencepiece}
Taku Kudo and John Richardson. 2018.
\newblock \href {https://doi.org/10.18653/v1/D18-2012} {{S}entence{P}iece: A simple and language independent subword tokenizer and detokenizer for neural text processing}.
\newblock In \emph{Proceedings of the 2018 Conference on Empirical Methods in Natural Language Processing: System Demonstrations}, pages 66--71, Brussels, Belgium. Association for Computational Linguistics.

\bibitem[{Lai et~al.(2017)Lai, Xie, Liu, Yang, and Hovy}]{lai-etal-2017-race}
Guokun Lai, Qizhe Xie, Hanxiao Liu, Yiming Yang, and Eduard Hovy. 2017.
\newblock \href {https://doi.org/10.18653/v1/D17-1082} {{RACE}: Large-scale {R}e{A}ding comprehension dataset from examinations}.
\newblock In \emph{Proceedings of the 2017 Conference on Empirical Methods in Natural Language Processing}, pages 785--794, Copenhagen, Denmark. Association for Computational Linguistics.

\bibitem[{Lan et~al.(2020)Lan, Chen, Goodman, Gimpel, Sharma, and Soricut}]{lan2019albert}
Zhenzhong Lan, Mingda Chen, Sebastian Goodman, Kevin Gimpel, Piyush Sharma, and Radu Soricut. 2020.
\newblock \href {https://openreview.net/forum?id=H1eA7AEtvS} {Albert: A lite bert for self-supervised learning of language representations}.
\newblock In \emph{International Conference on Learning Representations}.

\bibitem[{Li et~al.(2025)Li, Zheng, Wang, Zhang, Wang, Xuyang, Fan, Ding, Wang, Ding, Zhou, Zhang, and Jiang}]{li2025predictable}
Houyi Li, Wenzhen Zheng, Qiufeng Wang, Hanshan Zhang, Zili Wang, Shijie Xuyang, Yuantao Fan, Zhenyu Ding, Haoying Wang, Ning Ding, Shuigeng Zhou, Xiangyu Zhang, and Daxin Jiang. 2025.
\newblock \href {https://arxiv.org/abs/2503.04715} {Predictable scale: Part i, step law -- optimal hyperparameter scaling law in large language model pretraining}.
\newblock \emph{Preprint}, arXiv:2503.04715.

\bibitem[{Lin et~al.(2022)Lin, Hilton, and Evans}]{lin-etal-2022-truthfulqa}
Stephanie Lin, Jacob Hilton, and Owain Evans. 2022.
\newblock \href {https://doi.org/10.18653/v1/2022.acl-long.229} {{T}ruthful{QA}: Measuring how models mimic human falsehoods}.
\newblock In \emph{Proceedings of the 60th Annual Meeting of the Association for Computational Linguistics (Volume 1: Long Papers)}, pages 3214--3252, Dublin, Ireland. Association for Computational Linguistics.

\bibitem[{Mihaylov et~al.(2018)Mihaylov, Clark, Khot, and Sabharwal}]{mihaylov-etal-2018-suit}
Todor Mihaylov, Peter Clark, Tushar Khot, and Ashish Sabharwal. 2018.
\newblock \href {https://doi.org/10.18653/v1/D18-1260} {Can a suit of armor conduct electricity? a new dataset for open book question answering}.
\newblock In \emph{Proceedings of the 2018 Conference on Empirical Methods in Natural Language Processing}, pages 2381--2391, Brussels, Belgium. Association for Computational Linguistics.

\bibitem[{Minixhofer et~al.(2026)Minixhofer, Murray, Limisiewicz, Korhonen, Zettlemoyer, Smith, Ponti, Soldaini, and Hofmann}]{minixhofer2025bolmo}
Benjamin Minixhofer, Tyler Murray, Tomasz Limisiewicz, Anna Korhonen, Luke Zettlemoyer, Noah~A. Smith, Edoardo~M. Ponti, Luca Soldaini, and Valentin Hofmann. 2026.
\newblock \href {https://arxiv.org/abs/2512.15586} {Bolmo: Byteifying the next generation of language models}.
\newblock \emph{Preprint}, arXiv:2512.15586.

\bibitem[{Mostafazadeh et~al.(2016)Mostafazadeh, Chambers, He, Parikh, Batra, Vanderwende, Kohli, and Allen}]{mostafazadeh-etal-2016-corpus}
Nasrin Mostafazadeh, Nathanael Chambers, Xiaodong He, Devi Parikh, Dhruv Batra, Lucy Vanderwende, Pushmeet Kohli, and James Allen. 2016.
\newblock \href {https://doi.org/10.18653/v1/N16-1098} {A corpus and cloze evaluation for deeper understanding of commonsense stories}.
\newblock In \emph{Proceedings of the 2016 Conference of the North {A}merican Chapter of the Association for Computational Linguistics: Human Language Technologies}, pages 839--849, San Diego, California. Association for Computational Linguistics.

\bibitem[{Mundra et~al.(2024)Mundra, Khandavally, Dabre, Puduppully, Kunchukuttan, and Khapra}]{mundra-etal-2024-empirical}
Nandini Mundra, Aditya Nanda~Kishore Khandavally, Raj Dabre, Ratish Puduppully, Anoop Kunchukuttan, and Mitesh~M Khapra. 2024.
\newblock \href {https://doi.org/10.18653/v1/2024.conll-1.8} {An empirical comparison of vocabulary expansion and initialization approaches for language models}.
\newblock In \emph{Proceedings of the 28th Conference on Computational Natural Language Learning}, pages 84--104, Miami, FL, USA. Association for Computational Linguistics.

\bibitem[{Nakash et~al.(2025)Nakash, Calderon, Ben-David, Hoffer, and Reichart}]{nakash2025adaptivocab}
Itay Nakash, Nitay Calderon, Eyal Ben-David, Elad Hoffer, and Roi Reichart. 2025.
\newblock \href {https://openreview.net/forum?id=TyXf9dwpZP} {{AdaptiVocab}: Enhancing {LLM} efficiency in focused domains through lightweight vocabulary adaptation}.
\newblock In \emph{Second Conference on Language Modeling}.

\bibitem[{Owodunni and Kumar(2026)}]{owodunni2025continually}
Abraham~Toluwase Owodunni and Sachin Kumar. 2026.
\newblock \href {https://openreview.net/forum?id=HE84ER1BNL} {Continually adding new languages to multilingual language models}.
\newblock \emph{Transactions on Machine Learning Research}.

\bibitem[{Pagnoni et~al.(2025)Pagnoni, Pasunuru, Rodriguez, Nguyen, Muller, Li, Zhou, Yu, Weston, Zettlemoyer, Ghosh, Lewis, Holtzman, and Iyer}]{pagnoni-etal-2025-byte}
Artidoro Pagnoni, Ramakanth Pasunuru, Pedro Rodriguez, John Nguyen, Benjamin Muller, Margaret Li, Chunting Zhou, Lili Yu, Jason~E Weston, Luke Zettlemoyer, Gargi Ghosh, Mike Lewis, Ari Holtzman, and Srini Iyer. 2025.
\newblock \href {https://doi.org/10.18653/v1/2025.acl-long.453} {Byte latent transformer: Patches scale better than tokens}.
\newblock In \emph{Proceedings of the 63rd Annual Meeting of the Association for Computational Linguistics (Volume 1: Long Papers)}, pages 9238--9258, Vienna, Austria. Association for Computational Linguistics.

\bibitem[{Paperno et~al.(2016)Paperno, Kruszewski, Lazaridou, Pham, Bernardi, Pezzelle, Baroni, Boleda, and Fern{\'a}ndez}]{paperno-etal-2016-lambada}
Denis Paperno, Germ{\'a}n Kruszewski, Angeliki Lazaridou, Ngoc~Quan Pham, Raffaella Bernardi, Sandro Pezzelle, Marco Baroni, Gemma Boleda, and Raquel Fern{\'a}ndez. 2016.
\newblock \href {https://doi.org/10.18653/v1/P16-1144} {The {LAMBADA} dataset: Word prediction requiring a broad discourse context}.
\newblock In \emph{Proceedings of the 54th Annual Meeting of the Association for Computational Linguistics (Volume 1: Long Papers)}, pages 1525--1534, Berlin, Germany. Association for Computational Linguistics.

\bibitem[{Penedo et~al.(2025)Penedo, Kydl{\'\i}{\v{c}}ek, Sabol{\v{c}}ec, Messmer, Foroutan, Kargaran, Raffel, Jaggi, Werra, and Wolf}]{penedo2025fineweb2pipelinescale}
Guilherme Penedo, Hynek Kydl{\'\i}{\v{c}}ek, Vinko Sabol{\v{c}}ec, Bettina Messmer, Negar Foroutan, Amir~Hossein Kargaran, Colin Raffel, Martin Jaggi, Leandro~Von Werra, and Thomas Wolf. 2025.
\newblock \href {https://openreview.net/forum?id=jnRBe6zatP} {Fineweb2: One pipeline to scale them all {\textemdash} adapting pre-training data processing to every language}.
\newblock In \emph{Second Conference on Language Modeling}.

\bibitem[{Penedo et~al.(2024)Penedo, Kydl\'{\i}\v{c}ek, allal, Lozhkov, Mitchell, Raffel, Von~Werra, and Wolf}]{penedo2024fineweb}
Guilherme Penedo, Hynek Kydl\'{\i}\v{c}ek, Loubna~Ben allal, Anton Lozhkov, Margaret Mitchell, Colin Raffel, Leandro Von~Werra, and Thomas Wolf. 2024.
\newblock \href {https://doi.org/10.52202/079017-0970} {The {FineWeb} datasets: Decanting the web for the finest text data at scale}.
\newblock In \emph{Advances in Neural Information Processing Systems}, volume~37, pages 30811--30849. Curran Associates, Inc.

\bibitem[{Pilehvar et~al.(2018)Pilehvar, Kartsaklis, Prokhorov, and Collier}]{pilehvar-etal-2018-card}
Mohammad~Taher Pilehvar, Dimitri Kartsaklis, Victor Prokhorov, and Nigel Collier. 2018.
\newblock \href {https://doi.org/10.18653/v1/D18-1169} {Card-660: {C}ambridge rare word dataset - a reliable benchmark for infrequent word representation models}.
\newblock In \emph{Proceedings of the 2018 Conference on Empirical Methods in Natural Language Processing}, pages 1391--1401, Brussels, Belgium. Association for Computational Linguistics.

\bibitem[{Prakash et~al.(2020)Prakash, Shashidhar, Zhao, Rongali, Khan, and Kayser}]{prakash-etal-2020-compressing}
Prafull Prakash, Saurabh~Kumar Shashidhar, Wenlong Zhao, Subendhu Rongali, Haidar Khan, and Michael Kayser. 2020.
\newblock \href {https://doi.org/10.18653/v1/2020.findings-emnlp.423} {Compressing transformer-based semantic parsing models using compositional code embeddings}.
\newblock In \emph{Findings of the Association for Computational Linguistics: EMNLP 2020}, pages 4711--4717, Online. Association for Computational Linguistics.

\bibitem[{Press and Wolf(2017)}]{press-wolf-2017-using}
Ofir Press and Lior Wolf. 2017.
\newblock \href {https://aclanthology.org/E17-2025/} {Using the output embedding to improve language models}.
\newblock In \emph{Proceedings of the 15th Conference of the {E}uropean Chapter of the Association for Computational Linguistics: Volume 2, Short Papers}, pages 157--163, Valencia, Spain. Association for Computational Linguistics.

\bibitem[{Remy et~al.(2024)Remy, Delobelle, Avetisyan, Khabibullina, de~Lhoneux, and Demeester}]{remy2024transtokenization}
Fran{\c{c}}ois Remy, Pieter Delobelle, Hayastan Avetisyan, Alfiya Khabibullina, Miryam de~Lhoneux, and Thomas Demeester. 2024.
\newblock \href {https://openreview.net/forum?id=sBxvoDhvao} {Trans-tokenization and cross-lingual vocabulary transfers: Language adaptation of {LLM}s for low-resource {NLP}}.
\newblock In \emph{First Conference on Language Modeling}.

\bibitem[{Sankar et~al.(2021)Sankar, Ravi, and Kozareva}]{sankar-etal-2021-proformer}
Chinnadhurai Sankar, Sujith Ravi, and Zornitsa Kozareva. 2021.
\newblock \href {https://doi.org/10.18653/v1/2021.eacl-main.246} {{P}ro{F}ormer: Towards on-device {LSH} projection based transformers}.
\newblock In \emph{Proceedings of the 16th Conference of the European Chapter of the Association for Computational Linguistics: Main Volume}, pages 2823--2828, Online. Association for Computational Linguistics.

\bibitem[{Sap et~al.(2019)Sap, Rashkin, Chen, Le~Bras, and Choi}]{sap-etal-2019-social}
Maarten Sap, Hannah Rashkin, Derek Chen, Ronan Le~Bras, and Yejin Choi. 2019.
\newblock \href {https://doi.org/10.18653/v1/D19-1454} {Social {IQ}a: Commonsense reasoning about social interactions}.
\newblock In \emph{Proceedings of the 2019 Conference on Empirical Methods in Natural Language Processing and the 9th International Joint Conference on Natural Language Processing (EMNLP-IJCNLP)}, pages 4463--4473, Hong Kong, China. Association for Computational Linguistics.

\bibitem[{Sennrich et~al.(2016)Sennrich, Haddow, and Birch}]{sennrich-etal-2016-neural}
Rico Sennrich, Barry Haddow, and Alexandra Birch. 2016.
\newblock \href {https://doi.org/10.18653/v1/P16-1162} {Neural machine translation of rare words with subword units}.
\newblock In \emph{Proceedings of the 54th Annual Meeting of the Association for Computational Linguistics (Volume 1: Long Papers)}, pages 1715--1725, Berlin, Germany. Association for Computational Linguistics.

\bibitem[{Senuma(2025)}]{senuma2025mmh3}
Hajime Senuma. 2025.
\newblock \href {https://doi.org/10.21105/joss.06124} {mmh3: A python extension for murmurhash3}.
\newblock \emph{Journal of Open Source Software}, 10(105):6124.

\bibitem[{Shu and Nakayama(2018)}]{shu2018compressing}
Raphael Shu and Hideki Nakayama. 2018.
\newblock \href {https://openreview.net/forum?id=BJRZzFlRb} {Compressing word embeddings via deep compositional code learning}.
\newblock In \emph{International Conference on Learning Representations}.

\bibitem[{Skarda and Freeman(1987)}]{skarda1987brains}
Christine~A. Skarda and Walter~J. Freeman. 1987.
\newblock \href {https://doi.org/10.1017/S0140525X00047336} {How brains make chaos in order to make sense of the world}.
\newblock \emph{Behavioral and Brain Sciences}, 10(2):161–173.

\bibitem[{Socher et~al.(2011)Socher, Lin, Ng, and Manning}]{socher2011parsing}
Richard Socher, Cliff Chiung-Yu Lin, Andrew~Y. Ng, and Christopher~D. Manning. 2011.
\newblock Parsing natural scenes and natural language with recursive neural networks.
\newblock In \emph{Proceedings of the 28th International Conference on International Conference on Machine Learning}, ICML'11, page 129–136, Madison, WI, USA. Omnipress.

\bibitem[{{Team Olmo} et~al.(2026){Team Olmo}, Ettinger, Bertsch, Kuehl, Graham, Heineman, Groeneveld, Brahman, Timbers, Ivison, Morrison, Poznanski, Lo, Soldaini, Jordan, Chen, Noukhovitch, Lambert, Walsh, Dasigi, Berry, Malik, Shah, Geng, Arora, Gupta, Anderson, Xiao, Murray, Romero, Graf, Asai, Bhagia, Wettig, Liu, Rangapur, Anastasiades, Huang, Schwenk, Trivedi, Magnusson, Lochner, Liu, Miranda, Sap, Morgan, Schmitz, Guerquin, Wilson, Huff, Bras, Xin, Shao, Skjonsberg, Shen, Li, Wilde, Pyatkin, Merrill, Chang, Gu, Zeng, Sabharwal, Zettlemoyer, Koh, Farhadi, Smith, and Hajishirzi}]{olmo2026olmo3}
{Team Olmo}, Allyson Ettinger, Amanda Bertsch, Bailey Kuehl, David Graham, David Heineman, Dirk Groeneveld, Faeze Brahman, Finbarr Timbers, Hamish Ivison, Jacob Morrison, Jake Poznanski, Kyle Lo, Luca Soldaini, Matt Jordan, Mayee Chen, Michael Noukhovitch, Nathan Lambert, Pete Walsh, and 49 others. 2026.
\newblock \href {https://arxiv.org/abs/2512.13961} {Olmo 3}.
\newblock \emph{Preprint}, arXiv:2512.13961.

\bibitem[{Tejaswi et~al.(2024)Tejaswi, Gupta, and Choi}]{tejaswi-etal-2024-exploring}
Atula Tejaswi, Nilesh Gupta, and Eunsol Choi. 2024.
\newblock \href {https://doi.org/10.18653/v1/2024.findings-emnlp.614} {Exploring design choices for building language-specific {LLM}s}.
\newblock In \emph{Findings of the Association for Computational Linguistics: EMNLP 2024}, pages 10485--10500, Miami, Florida, USA. Association for Computational Linguistics.

\bibitem[{Tito~Svenstrup et~al.(2017)Tito~Svenstrup, Hansen, and Winther}]{svenstrup2017hash}
Dan Tito~Svenstrup, Jonas Hansen, and Ole Winther. 2017.
\newblock \href {https://proceedings.neurips.cc/paper_files/paper/2017/file/f0f6ba4b5e0000340312d33c212c3ae8-Paper.pdf} {Hash embeddings for efficient word representations}.
\newblock In \emph{Advances in Neural Information Processing Systems}, volume~30. Curran Associates, Inc.

\bibitem[{Tsuda(2001)}]{tsuda2001toward}
Ichiro Tsuda. 2001.
\newblock \href {https://doi.org/10.1017/S0140525X01000097} {Toward an interpretation of dynamic neural activity in terms of chaotic dynamical systems}.
\newblock \emph{Behavioral and Brain Sciences}, 24(5):793–810.

\bibitem[{Vaswani et~al.(2017)Vaswani, Shazeer, Parmar, Uszkoreit, Jones, Gomez, Kaiser, and Polosukhin}]{vaswani2017attention}
Ashish Vaswani, Noam Shazeer, Niki Parmar, Jakob Uszkoreit, Llion Jones, Aidan~N Gomez, \L~ukasz Kaiser, and Illia Polosukhin. 2017.
\newblock \href {https://proceedings.neurips.cc/paper_files/paper/2017/file/3f5ee243547dee91fbd053c1c4a845aa-Paper.pdf} {Attention is all you need}.
\newblock In \emph{Advances in Neural Information Processing Systems}, volume~30. Curran Associates, Inc.

\bibitem[{Videau et~al.(2025)Videau, Youbi~Idrissi, Leite, Schoenauer, Teytaud, and Lopez-Paz}]{videau2026bytes}
Mathurin Videau, Badr Youbi~Idrissi, Alessandro Leite, Marc Schoenauer, Olivier Teytaud, and David Lopez-Paz. 2025.
\newblock \href {https://doi.org/10.52202/085713-3144} {From bytes to ideas: Language modeling with autoregressive {U-Net}s}.
\newblock In \emph{Advances in Neural Information Processing Systems}, volume 38, Main Conference, pages 93991--94009. Curran Associates, Inc.

\bibitem[{Welbl et~al.(2017)Welbl, Liu, and Gardner}]{welbl-etal-2017-crowdsourcing}
Johannes Welbl, Nelson~F. Liu, and Matt Gardner. 2017.
\newblock \href {https://doi.org/10.18653/v1/W17-4413} {Crowdsourcing multiple choice science questions}.
\newblock In \emph{Proceedings of the 3rd Workshop on Noisy User-generated Text}, pages 94--106, Copenhagen, Denmark. Association for Computational Linguistics.

\bibitem[{Xue and Aletras(2022)}]{xue-aletras-2022-hashformers}
Huiyin Xue and Nikolaos Aletras. 2022.
\newblock \href {https://doi.org/10.18653/v1/2022.emnlp-main.536} {{H}ash{F}ormers: Towards vocabulary-independent pre-trained transformers}.
\newblock In \emph{Proceedings of the 2022 Conference on Empirical Methods in Natural Language Processing}, pages 7862--7874, Abu Dhabi, United Arab Emirates. Association for Computational Linguistics.

\bibitem[{Xue et~al.(2025)Xue, Moosavi, and Aletras}]{xue-etal-2025-deconstructing}
Huiyin Xue, Nafise~Sadat Moosavi, and Nikolaos Aletras. 2025.
\newblock \href {https://doi.org/10.18653/v1/2025.ijcnlp-long.40} {Deconstructing attention: Investigating design principles for effective language modeling}.
\newblock In \emph{Proceedings of the 14th International Joint Conference on Natural Language Processing and the 4th Conference of the Asia-Pacific Chapter of the Association for Computational Linguistics}, pages 708--727, Mumbai, India. The Asian Federation of Natural Language Processing and The Association for Computational Linguistics.

\bibitem[{Xue et~al.(2022)Xue, Barua, Constant, Al-Rfou, Narang, Kale, Roberts, and Raffel}]{xue-etal-2022-byt5}
Linting Xue, Aditya Barua, Noah Constant, Rami Al-Rfou, Sharan Narang, Mihir Kale, Adam Roberts, and Colin Raffel. 2022.
\newblock \href {https://doi.org/10.1162/tacl_a_00461} {{B}y{T}5: Towards a token-free future with pre-trained byte-to-byte models}.
\newblock \emph{Transactions of the Association for Computational Linguistics}, 10:291--306.

\bibitem[{Yamaguchi et~al.(2026{\natexlab{a}})Yamaguchi, Mi, and Aletras}]{yamaguchi-etal-2026-enhancing}
Atsuki Yamaguchi, Maggie Mi, and Nikolaos Aletras. 2026{\natexlab{a}}.
\newblock \href {https://doi.org/10.18653/v1/2026.acl-short.27} {Enhancing linguistic competence of language models through pre-training with language learning tasks}.
\newblock In \emph{Proceedings of the 64th Annual Meeting of the {A}ssociation for {C}omputational {L}inguistics (Volume 2: Short Papers)}, pages 316--336, San Diego, California, United States. Association for Computational Linguistics.

\bibitem[{Yamaguchi et~al.(2025)Yamaguchi, Morishita, Villavicencio, and Aletras}]{yamaguchi2025adapting}
Atsuki Yamaguchi, Terufumi Morishita, Aline Villavicencio, and Nikolaos Aletras. 2025.
\newblock \href {https://openreview.net/forum?id=6IdoIKowfe} {Adapting chat language models using only target unlabeled language data}.
\newblock \emph{Transactions on Machine Learning Research}.

\bibitem[{Yamaguchi et~al.(2026{\natexlab{b}})Yamaguchi, Villavicencio, and Aletras}]{10.1162/COLI.a.581}
Atsuki Yamaguchi, Aline Villavicencio, and Nikolaos Aletras. 2026{\natexlab{b}}.
\newblock \href {https://doi.org/10.1162/COLI.a.581} {How can we effectively expand the vocabulary of {LLM}s with 0.01{GB} of target language text?}
\newblock \emph{Computational Linguistics}, 52(1):295--330.

\bibitem[{Yang et~al.(2025)Yang, Li, Yang, Zhang, Hui, Zheng, Yu, Gao, Huang, Lv, Zheng, Liu, Zhou, Huang, Hu, Ge, Wei, Lin, Tang, Yang, Tu, Zhang, Yang, Yang, Zhou, Zhou, Lin, Dang, Bao, Yang, Yu, Deng, Li, Xue, Li, Zhang, Wang, Zhu, Men, Gao, Liu, Luo, Li, Tang, Yin, Ren, Wang, Zhang, Ren, Fan, Su, Zhang, Zhang, Wan, Liu, Wang, Cui, Zhang, Zhou, and Qiu}]{yang2025qwen3technicalreport}
An~Yang, Anfeng Li, Baosong Yang, Beichen Zhang, Binyuan Hui, Bo~Zheng, Bowen Yu, Chang Gao, Chengen Huang, Chenxu Lv, Chujie Zheng, Dayiheng Liu, Fan Zhou, Fei Huang, Feng Hu, Hao Ge, Haoran Wei, Huan Lin, Jialong Tang, and 41 others. 2025.
\newblock \href {https://arxiv.org/abs/2505.09388} {Qwen3 technical report}.
\newblock \emph{Preprint}, arXiv:2505.09388.

\bibitem[{Yang et~al.(2018)Yang, Dai, Salakhutdinov, and Cohen}]{yang2017breaking}
Zhilin Yang, Zihang Dai, Ruslan Salakhutdinov, and William~W. Cohen. 2018.
\newblock \href {https://openreview.net/forum?id=HkwZSG-CZ} {Breaking the softmax bottleneck: A high-rank {RNN} language model}.
\newblock In \emph{International Conference on Learning Representations}.

\bibitem[{Yao et~al.(2021)Yao, Huang, Wang, Dong, and Wei}]{yao-etal-2021-adapt}
Yunzhi Yao, Shaohan Huang, Wenhui Wang, Li~Dong, and Furu Wei. 2021.
\newblock \href {https://doi.org/10.18653/v1/2021.findings-acl.40} {Adapt-and-distill: Developing small, fast and effective pretrained language models for domains}.
\newblock In \emph{Findings of the Association for Computational Linguistics: ACL-IJCNLP 2021}, pages 460--470, Online. Association for Computational Linguistics.

\bibitem[{Zellers et~al.(2019)Zellers, Holtzman, Bisk, Farhadi, and Choi}]{zellers-etal-2019-hellaswag}
Rowan Zellers, Ari Holtzman, Yonatan Bisk, Ali Farhadi, and Yejin Choi. 2019.
\newblock \href {https://doi.org/10.18653/v1/P19-1472} {{H}ella{S}wag: Can a machine really finish your sentence?}
\newblock In \emph{Proceedings of the 57th Annual Meeting of the Association for Computational Linguistics}, pages 4791--4800, Florence, Italy. Association for Computational Linguistics.

\bibitem[{Zhang et~al.(2018)Zhang, Liu, Liu, Gao, Duh, and Durme}]{zhang2018record}
Sheng Zhang, Xiaodong Liu, Jingjing Liu, Jianfeng Gao, Kevin Duh, and Benjamin~Van Durme. 2018.
\newblock \href {https://arxiv.org/abs/1810.12885} {Record: Bridging the gap between human and machine commonsense reading comprehension}.
\newblock \emph{Preprint}, arXiv:1810.12885.

\end{thebibliography}

\clearpage

\appendix

\section{Softmax Bottleneck}\label{appendix:softmax_bottleneck}

\subsection{Preliminaries}
Language models (LMs) define the conditional distribution $P_\theta(w|c)$ by applying a softmax function to a linear projection of the hidden state $\mathbf{h}_t$:

{\small
\begin{align*}
P_\theta(w|c)=\frac{\text{exp }\mathbf{h}_t^\top\mathbf{w}_{w}}{\sum_{w^\prime}\text{exp }\mathbf{h}_t^\top\mathbf{w}_{w^\prime}}
\end{align*}
}

\noindent where both the context vector $\mathbf{h}_t(c_t; \theta)$ and the word embedding $\mathbf{w}_{w_t}(w_t; \theta)$ are on a $d$-dimensional space. Their inner product, $\mathbf{h}_t^\top \mathbf{w}_{w_t}$, defines the logit. To analyze the expressive capacity of the softmax function, we define the following three matrices and their log-probabilities:

{\small
\begin{align*}
\mathbf{H}_\theta = [\mathbf{h}_{1}, \dots, \mathbf{h}_{n}]^\top, \quad \mathbf{W}_\theta = [\mathbf{w}_{1}, \dots, \mathbf{w}_{\vert\mathcal{V}\vert}]^\top \\
\mathbf{A} = [a_{i,j}]_{n \times {\vert\mathcal{V}\vert}} \in \mathbb{R}^{n \times {\vert\mathcal{V}\vert}}, \text{ where } a_{i,j} = \log p^\ast(w_j|c_i)
\end{align*}
}

{\small
\begin{align*}
    \log p^\ast(w|c) &= \log \left( \frac{\exp(\mathbf{h}_t^\top\mathbf{w}_w)}{\sum_{w^\prime} \exp(\mathbf{h}_t^\top\mathbf{w}_{w^\prime})} \right) \\
    &= \mathbf{h}_t^\top\mathbf{w}_w - \log \sum_{w^\prime} \exp(\mathbf{h}_t^\top\mathbf{w}_{w^\prime})
\end{align*}
}

\noindent$\mathbf{H}_\theta \in \mathbb{R}^{n \times d}$ denote the matrix of $\mathbf{h}_t$ and $\mathbf{W}_\theta \in \mathbb{R}^{\vert\mathcal{V}\vert \times d}$ is the projection matrix for all tokens in the vocabulary. The matrix $\mathbf{A} \in \mathbb{R}^{n \times {\vert\mathcal{V}\vert}}$ contains the log probabilities of the true data distribution. We index $\mathbf{H}$ and $\mathbf{W}$ by $\theta$ to indicate that both represent functions within a joint family $\mathcal{U}$ parameterized by $\theta$. In practice, $\mathbf{H}_\theta$ is realized via a Transformer-based LM, while $\mathbf{W}_\theta$ is instantiated as a learned token embedding lookup table.

We further define a set of matrices, $F(\mathbf{A})$, generated by applying a row-wise shift to $\mathbf{A}$:
{\small
\begin{align*}
F(\mathbf{A}) = \{ \mathbf{A} + \Lambda \mathbf{J}_{n,{\vert\mathcal{V}\vert}} \mid \Lambda \in \mathbb{R}^{n \times n} \text{ is a diagonal matrix} \}
\end{align*}
}

\noindent where $\mathbf{J}_{n,{\vert\mathcal{V}\vert}} \in \mathbb{R}^{n \times {\vert\mathcal{V}\vert}}$ denotes the all-ones matrix. This row-wise shift operation effectively adds a unique scalar constant to each element of $\mathbf{A}$'s rows. Since the diagonal entries of $\Lambda$ can be any real values, $F(\mathbf{A})$ constitutes an infinite set characterized by two key properties:

\begin{property}
For any matrix $\mathbf{A}^\prime$, $\mathbf{A}^\prime \in F(\mathbf{A})$ if and only if $\text{Softmax}(\mathbf{A}^\prime) = P^\ast$. Thus, $F(\mathbf{A})$ represents the set of all logit matrices that recover the true data distribution.
\end{property}\label{property:property_1}

\begin{property}
For any $\mathbf{A}_1, \mathbf{A}_2 \in F(\mathbf{A})$ such that $\mathbf{A}_1 \neq \mathbf{A}_2$, it holds that $|\text{rank}(\mathbf{A}_1) - \text{rank}(\mathbf{A}_2)| \leq 1$. Hence, the matrices in $F(\mathbf{A})$ maintain consistent ranks, with a maximum discrepancy of one.
\end{property}\label{property:property_2}

Given Property~\ref{property:property_1}, we establish: 
\begin{lemma}
Given a model parameter $\theta$, $\mathbf{H}_\theta\mathbf{W}_\theta^\top \in F(\mathbf{A})$ if and only if $P_\theta(X|c)$ holds for all $c \in \mathcal{L}$.
\end{lemma}

Following this lemma, expressiveness is framed as follows: does there exist a parameter $\theta$ and a matrix $\mathbf{A}^\prime \in F(\mathbf{A})$ satisfying
$\mathbf{H}_\theta\mathbf{W}_\theta^\top = \mathbf{A}^\prime$. LMs learn matrices $\mathbf{H}_\theta$ and $\mathbf{W}_\theta$ to factorize a target matrix $\mathbf{A}^\prime \in F(\mathbf{A})$. For a valid factorization, the rank of the product $\mathbf{H}_\theta\mathbf{W}_\theta^\top$ must be at least equal to the rank of $\mathbf{A}^\prime$. Because $\mathbf{H}_\theta \in \mathbb{R}^{n \times d}$ and $\mathbf{W}_\theta \in \mathbb{R}^{{\vert\mathcal{V}\vert} \times d}$, the rank of this product is strictly upper-bounded by the embedding dimension $d$. Consequently, if $d \geq \text{rank}(\mathbf{A}^\prime)$, a universal approximator can recover $\mathbf{A}^\prime$. Conversely, if $d < \text{rank}(\mathbf{A}^\prime)$, no pair $(\mathbf{H}_\theta, \mathbf{W}_\theta)$ can recover $\mathbf{A}^\prime$, regardless of the expressiveness of $\mathcal{U}$.

\begin{proposition} 
Assuming the function family $\mathcal{U}$ is a universal approximator, there exists a parameter $\theta$ such that $P_\theta(X|c) = P^\ast(X|c)$ for all $c \in \mathcal{L}$ if and only if $d \geq \min_{\mathbf{A}^\prime \in F(\mathbf{A})} \text{rank}(\mathbf{A}^\prime)$.
\end{proposition}\label{proposition:proposition_1}

\noindent By combining Proposition~\ref{proposition:proposition_1} with the properties of $F(\mathbf{A})$ outlined in Property~\ref{property:property_2}, \citet{yang2017breaking} define the Softmax Bottleneck:

\begin{corollary}
\textbf{\textit{(Softmax Bottleneck)}}. If $d < \text{rank}(\mathbf{A}) - 1$, then for any function family $\mathcal{U}$ that acts as a universal approximator, there is no parameter $\theta$ such that $P_\theta(X|c) = P^\ast(X|c)$ for all $c \in \mathcal{L}$. That is, $P_\theta(X|c) \neq P^\ast(X|c)$ must hold for at least some contexts.
\end{corollary}

\noindent This demonstrates that an insufficient dimension $d$ restricts Softmax from expressing the true data distribution. The conclusion is not restricted to a finite language $\mathcal{L}$. Because when $\mathcal{L}$ is infinite, one can always take a finite subset and the softmax bottleneck still exists.

\subsection{\textsc{MultiHashFormer} as a Mitigation Strategy}
While training, \textsc{MultiHashFormer} approximates elements in $\mathbf{A}$ by
{\small
\begin{align*}
    \log p^\ast(w|c) &= \log \prod_{i}^{H} p^\ast(\mathcal{H}_i(w)|c)\\
    &= \sum_{i=1}^{H}\log p^\ast(\mathcal{H}_i(w)|c)\\
    &= \sum_{i=1}^{H}\log \left( \frac{\exp(\mathbf{c}^{(i)\top}\mathbf{w}^{(i)}_{\mathcal{H}_i(w)})\pi^{(i)}_{\mathcal{H}_i(w)}}{\sum_{j=1}^B \exp(\mathbf{c}^{(i)\top}\mathbf{w}^{(i)})} \right) \\
    &= \sum_{i=1}^{H}\mathbf{c}^{(i)\top}\mathbf{w}^{(i)}_{\mathcal{H}_i(w)}\pi^{(i)}_{\mathcal{H}_i(w)} \\ 
    &\quad- \sum_{i=1}^{H}\log \sum_{j=1}^{B} \exp(\mathbf{c}^{(i)\top}\mathbf{w}^{(i)}_{j})
\end{align*}
}

\noindent The model projects representations to coordinates of the hash signature, using $H$ localized parameter weight matrices assigned exclusively to $i$-th coordinate bucket array, $\mathbf{W}_o^{(i)} \in \mathbb{R}^{d \times B}$. These coordinates are then mapped to the true vocabulary space via transition matrices $\pi^{(i)} \in \{0, 1\}^{B \times |\mathcal{V}|}$. Here, each column of $\pi^{(i)}$ is a one-hot vector, satisfying $\sum_{j=1}^{B} \pi^{(i)}_{j,k} = 1$ for every column $k$.

\noindent The estimation could be further written in a matrix form:

{\small
\begin{align*}
\hat{\mathbf{A}}_{MHF} &= \sum_{i=1}^{H}\mathbf{H}_\theta^{(i)}\mathbf{W}_o^{(i)\top}\pi^{(i)}\\
&\quad-\sum_{i=1}^{H}\log\left(\sum_{j=1}^{B}\exp(\mathbf{H}_\theta^{(i)}\mathbf{W}_o^{(i)\top})\right)
\end{align*}
}

\noindent $\mathbf{H}_\theta^{i}$ are not column-equivalent due to the softmax retrieval for the cascade mixer. Meanwhile, the residual connection in its state update can prevent rank collapse even with a low bottleneck projection dimension in the cascade mixer. The effectiveness of this design is also corroborated in \citet{hu2021lora} and \citet{xue-etal-2025-deconstructing}, where a low-rank projection in the weight update and deconstructed attention respectively.  Consequently, $\text{rank}(\mathbf{\hat{A}}_{\text{MHF}}) \leq \min(n,B, \vert\mathcal{V}_{actl.}\vert$, $H \times \min(d, B))= \min(B,H\times d$), where $H$ denotes the number of hash functions configured by \textsc{MultiHashFormer}, $B$ denotes the number of discrete hash buckets per function, and $d$ is the dimensionality of the latent hash state vectors $\mathbf{c}^{(i)}$. Given that the rank of Standard baselines is bounded by $\text{rank}(\mathbf{\hat{A}}_{\text{Standard}}) \leq \min(d, n, \vert\mathcal{V}_{\text{actl.}}\vert) = d$, \textsc{MultiHashFormer} elevates the upper bound of this rank to enhance model expressiveness, while employing multi hash functions and non-linear transformation between latent hash state vectors. This aligns with the intuition that aggregating multiple distinct observations inherently improves estimation accuracy.

During inference, although we exclude all unassigned signatures, the rank upper bound remains invariant between training and inference. This stability stems from the fact that the row-wise partition function does not contribute to the final upper bound of the rank. While MultiHashFormer does not produce an arbitrary rank, it elevates the rank of the conditional distribution matrix. This improvement drives performance gains in high-entropy reasoning tasks, including LAMBADA and HellaSwag.

\section{Rank-preserving Compositional Embedding}
Typical factorized embeddings~\citep{lan2019albert} use a low-rank matrix factorization across the hidden dimension. This inherently imposes a strict rank bottleneck on the representational capacity of the model. \textsc{MultiHashFormer} factorizes only across the vocabulary dimension via combinatorial coordinates. This avoids a simple low-dimensional intermediate matrix product along the lowest deterministic dimension (i.e., the hidden dimension), and leverages gated compositional layer to reconstruct the token embedding without collapsing the dimensional rank. The strong empirical results on zero parameter vocabulary expansion confirm this theoretical grounding.

\section{Rare Word Examples from Card-660}\label{appendix:card660_appx}

\begin{table}[!t]
\centering
\resizebox{1.0\linewidth}{!}{
\begin{NiceTabular}{lccccc}
\toprule
\textbf{Word Pair} & \textbf{Human} & \textbf{Stnd.} & \textbf{H3B10K} & \textbf{Stnd.+2L} & \textbf{H4B16K}\\
\midrule
\\
\multicolumn{6}{c}{\textbf{Abbrevation}}\\
\multicolumn{6}{l}{retweeting}\\ $\leftrightarrow$ RTing & 1.00 & 0.56 & 0.65 & 0.47 & 0.71\\
\multicolumn{6}{l}{science-fiction}\\	$\leftrightarrow$ sci-fi & 1.00 & 0.77 & 0.90 & 0.77 & 0.86\\
\multicolumn{6}{l}{Brooklyn}\\ $\leftrightarrow$ Brklyn & 1.00 & 0.45 & 0.53 & 0.46 & 0.61\\
\multicolumn{6}{l}{Internet Explorer 4}\\	$\leftrightarrow$  IE4 & 1.00 & 0.51 & 0.61 & 0.53 & 0.61\\
\multicolumn{6}{l}{ITV2}\\ $\leftrightarrow$ ITV Two & 1.00 & 0.51 & 0.60 & 0.43 & 0.52\\
\multicolumn{6}{l}{electrolytic polishing}\\	$\leftrightarrow$ electropolishing & 0.99 & 0.74 & 0.83 & 0.62 & 0.87\\
\\
\multicolumn{6}{c}{\textbf{Alias}}\\
\multicolumn{6}{l}{full-HD}\\ $\leftrightarrow$ 1080p & 1.00 & 0.55 & 0.72 & 0.40 & 0.67\\
\multicolumn{6}{l}{Malva parviflora}\\ $\leftrightarrow$ cheeseweed & 1.00 & 0.52 & 0.58 & 0.57 & 0.61\\
\multicolumn{6}{l}{first milk}\\ $\leftrightarrow$ colostrum & 1.00 & 0.54 & 0.64 & 0.53 & 0.69\\
\multicolumn{6}{l}{little bee-eater}\\ $\leftrightarrow$  Merops pusillus & 1.00 & 0.48 &	0.60 & 0.51 & 0.64\\
\\
\multicolumn{6}{c}{\textbf{Misspelling}}\\
\multicolumn{6}{l}{leggin}\\ $\leftrightarrow$ legging & 1.00 & 0.59 & 0.57 & 0.39 & 0.60\\
\multicolumn{6}{l}{tariqa}\\ $\leftrightarrow$ tariqah & 1.00 & 0.77 & 0.82 &	0.72 & 0.82\\
\multicolumn{6}{l}{shit}\\ $\leftrightarrow$ shxt & 0.99 & 0.17 &	0.53 & 0.22	& 0.52\\
\multicolumn{6}{l}{sweeeet}\\ $\leftrightarrow$ sweet & 0.99 & 0.36 &	0.45 & 0.37 & 0.52\\
\\
\multicolumn{6}{c}{\textbf{Synonym}}\\
\multicolumn{6}{l}{25}\\ $\leftrightarrow$ twenty-five & 1.00 & 0.75 & 0.81 & 0.75 & 0.84\\
\multicolumn{6}{l}{shapeless}\\ $\leftrightarrow$ amorphous & 0.99 & 0.38 & 0.50 & 0.51 & 0.63\\
\multicolumn{6}{l}{unforeseen}\\ $\leftrightarrow$ unanticipated & 0.97 & 0.82 &	0.91 &	0.83 & 0.93\\
\bottomrule

\end{NiceTabular}
}
\caption{Examples of word pairs from Card-660 with high human-annotated similarity scores (normalized to [0, 1]), compared against cosine similarities computed from the second-to-last decoder layer of the 1B Standard baselines and \textsc{MultiHashFormer} models. 
}
\label{table:card660_examples}

\end{table}
 
We first look to word pairs with high human-annotated similarity to serve as hard examples for our illustration. Table~\ref{table:card660_examples} presents examples of high-similarity word pairs from the Card-660 dataset. We compare normalized human annotation scores against cosine similarities derived from the second-to-last decoder layers of the 1B Standard baseline and \textsc{MultiHashFormer}. Compared to the Standard baseline, \textsc{MultiHashFormer} consistently assigns higher similarity scores to semantically equivalent word pairs, such as abbreviations, aliases, misspellings, and synonyms. Examples from word pairs with low similarities and medium similarities are shown in Table~\ref{table:card660_examples_other}.

\begin{table}[!t]
\centering
\resizebox{1.0\linewidth}{!}{
\begin{NiceTabular}{lccccc}
\toprule
\textbf{Word Pair} & \textbf{Human} & \textbf{Stnd.} & \textbf{H3B10K} & \textbf{Stnd.+2L} & \textbf{H4B16K}\\
\midrule
\\
\multicolumn{6}{c}{\textbf{Low}}\\
\multicolumn{6}{l}{NetMeeting}\\ $\leftrightarrow$ Marwar Hall & 0.00 & 0.37 & 0.41 & 0.39	& 0.34\\
\multicolumn{6}{l}{Park Ji-sung}\\ $\leftrightarrow$ Yosemite Park & 0.00 & 0.36 & 0.49 & 0.36 & 0.43\\
\multicolumn{6}{l}{Pizza Hut}\\ $\leftrightarrow$ Pizzle rot & 0.02 & 0.29 & 0.25	& 0.31 & 0.27\\
\multicolumn{6}{l}{cheddah}\\ $\leftrightarrow$ cheddar & 0.06 & 0.44 &	0.23 & 0.47	& 0.51\\
\multicolumn{6}{l}{cheddah}\\ $\leftrightarrow$ cheddar & 0.06 & 0.44 &	0.23 & 0.47	& 0.51\\
\multicolumn{6}{l}{Apple}\\ $\leftrightarrow$ Applebees & 0.10 & 0.49 & 0.56 & 0.49 & 0.65\\
\\

\multicolumn{6}{c}{\textbf{Medium}}\\
\multicolumn{6}{l}{Ben-Hur}\\ $\leftrightarrow$ Titanic & 0.39 & 0.40 & 0.39 & 0.27 & 0.50\\
\multicolumn{6}{l}{radionavigation}\\ $\leftrightarrow$ frequency band & 0.44 & 0.37 & 0.44 & 0.34	& 0.42\\
\multicolumn{6}{l}{night sky}\\ $\leftrightarrow$ skyglow & 0.49 & 0.60 & 0.65 & 0.60 & 0.64\\
\multicolumn{6}{l}{circus}\\ $\leftrightarrow$ ropedancer & 0.50 & 0.46 & 0.56 & 0.51 & 0.47\\
\multicolumn{6}{l}{transmigration}\\ $\leftrightarrow$ residence permit & 0.52 & 0.44 & 0.64 & 0.50 & 0.57\\
\multicolumn{6}{l}{Head tilt}\\ $\leftrightarrow$ cervix & 0.52 & 0.35 & 0.49 & 0.34 & 0.49\\

\bottomrule

\end{NiceTabular}
}
\caption{Examples of word pairs from Card-660 with low and medium human-annotated similarity scores (normalized to [0, 1]), compared against cosine similarities computed from the second-to-last decoder layer of the 1B Standard baselines and \textsc{MultiHashFormer} models. A human annotation score of 1.00 indicates semantic equivalence.
}
\label{table:card660_examples_other}

\end{table}

\newpage

\section{Hyperparameters}\label{appendix:hyperparam}

\subsection{Hyperparameters for Pre-training}

\begin{table}[!h]
\begin{center}
\resizebox{\columnwidth}{!}{%
\begin{tabularx}{\textwidth}{lXXX}
\toprule
\textbf{Hyperparameters} & \textbf{100M} & \textbf{1B} & \textbf{3B}\\
\midrule
Hidden size & 768 & 2048 & 2048\\
Intermediate size & 2560 & 6144 & 11008\\
Max window layers & 12 & 20 & 36\\
Added layers (if scaling the depth) & 4 & 2 & 1 \\
Number of attention heads & 12 & 16 & 16\\
Number of hidden layers & 12 & 20 & 36\\
Number of key value heads & 2 & 8 & 2\\
Rope theta & 1M & 1M & 1M\\
RMS norm eps & 1e-06 & 1e-06 & 1e-06\\
Attention dropout & 0.0 & 0.0 & 0.0\\
Tie word embeddings & True & True & True\\
Hidden activation & SiLU & SiLU & SiLU\\
Initializer range & 0.02 & 0.02 & 0.02\\
Vocabulary size & 32,768 & 32,768 & 32,768\\
Tokenizer & Mistral & Mistral & Mistral\\
Batch size & 256 & 256 & 256\\
Train steps  & 20K & 200K & 200K\\
Sequence length & 2,048 & 2,048 & 2,048\\
Maximum Learning Rate & 5e-4 & 3e-4 & 2e-4\\
Learning rate scheduler & cosine & cosine & cosine\\
Warmup steps & 2000 & 2000 & 2000\\
Optimizer & AdamW & AdamW & AdamW\\
Adam $\epsilon$ & 1e-8 & 1e-8 & 1e-8\\
Adam $\beta_1$ & 0.9 & 0.9 & 0.9\\
Adam $\beta_2$ & 0.999 & 0.999 & 0.999\\
Gradient clipping & 1.0 & 1.0 & 1.0\\
Weight decay & 0.1 & 0.1 & 0.1\\
Training precision & BF16 & BF16 & BF16\\
\bottomrule
\end{tabularx}%
}%
\caption{Hyperparameters of pre-training at each model scale.}
\label{table:hyperparams_pt}
\end{center}
\end{table}

\newpage

\subsection{Hyperparameters for VE Continual-pretraining}
\begin{table}[!h]
\begin{center}
\small
\begin{tabularx}{\linewidth}{lXX}
\toprule
\textbf{Hyperparameters} & \textbf{1B} & \textbf{3B}\\
\midrule
Adaptive Decoder Layer Indices & [1,2,19,20] & [1,2,35,36]\\
Attention dropout & 0.0 & 0.0\\
Tie word embeddings & True & True\\
Vocabulary size & 48,122 & 48,122\\
Tokenizer & expanded Mistral & expanded Mistral\\
Batch size & 256 & 256\\
Train steps  & 16K & 16K\\
Sequence length & 2,048 & 2,048\\
Maximum Learning Rate & 3e-4 & 2e-4\\
Learning rate scheduler & cosine & cosine\\
Warmup steps & 800 & 800\\
Optimizer & AdamW & AdamW\\
Adam $\epsilon$ & 1e-8 & 1e-8\\
Adam $\beta_1$ & 0.9 & 0.9\\
Adam $\beta_2$ & 0.999 & 0.999\\
Gradient clipping & 1.0 & 1.0\\
Weight decay & 0.1 & 0.1\\
Training precision & BF16 & BF16\\
\bottomrule
\end{tabularx}%
\caption{Additional hyperparameters of continual-pretraining at each model scale.}
\label{table:hyperparams_cpt}
\end{center}
\end{table}

\section{MultiHashFormer Abbrevations and Configurations}\label{appendix:MHF_abbr}

Table~\ref{table:abbr_mhf} presents the detailed configurations of \textsc{MultiHashFormer} variants  along with their abbreviations.

\begin{table}[h!]
\centering
\small
\resizebox{0.5\linewidth}{!}{
\begin{NiceTabular}{lcc}
\toprule
\textbf{MHF abbr.} & \textbf{H} & \textbf{B}\\
\midrule
H3B4K & 3 & 4,096\\
H3B8K & 3 & 8,192\\
H3B10K & 3 & 10,624\\
H3B16K & 3 & 16,384\\
H4B4K & 4 & 4,096\\
H4B8K & 4 & 8,192\\
H4B16K & 4 & 16,384\\
H4B32K & 4 & 32,768\\

\bottomrule
\end{NiceTabular}%
}
\caption{Detailed configurations of \textsc{MultiHashFormer} variants along with their abbreviations.}
\label{table:abbr_mhf}

\end{table}

\newpage

\section{Multiple-Training Runs at 100M Scale}
Regarding seed variance, we pre-train \textsc{MultiHashFormer} and the Standard baselines at 100M scale across 3 random seeds, averaging the scores over the same downstream tasks (see Table~\ref{table:performance_multi_run_100M}). \textsc{MultiHashFormer} variants outperform Standard and Standard+4L on 5 out of 10 tasks. These results are consistent with our single-run findings in Table~\ref{table:performance_1B_3B}.

\begin{table*}[!t]
\centering
\small
\resizebox{\linewidth}{!}{
\begin{NiceTabular}{clcccccccccc}
\toprule
& & \textbf{Lang. Model.} & \multicolumn{5}{c}{\textbf{Commonsense Reasoning}} & \multicolumn{3}{c}{\textbf{Reading Comprehension}} & \textbf{CR+RC}\\
\cmidrule(lr){3-3}   
\cmidrule(lr){4-8}   
\cmidrule(lr){9-11}  
\cmidrule(lr){11-11} 

& \textbf{Model} & \textbf{LAMBADA} & \textbf{ARC-E} & \textbf{COPA} & \textbf{OBQA} & \textbf{PIQA} & \textbf{HellaSwag} & \textbf{RACE} & \textbf{SciQ} & \textbf{SIQA} & \textbf{ReCoRD}\\
\midrule
& Rnd. Guess & - & 25.00 & 50.00 & 25.00 & 50.00 & 25.00 & 25.00 & 25.00 & 25.00 & 19.10\\
\midrule

\multirow{4}{*}{\rotatebox[origin=c]{90}{100M}} & Standard & 15.18 (0.4) & 41.49 (1.0)	& \textbf{61.00} (4.6) & 26.60 (2.0)	& 58.94 (1.0) & 28.32 (1.4)	& 25.90 (1.4) & 58.23 (1.6)	& 35.79 (1.0) & 49.05 (0.5)\\
& Standard+4L & \textbf{19.19} (0.6) & 44.98 (1.0) & 60.67 (4.4) & \textbf{27.47} (2.0) & 60.74 (1.1) & \textbf{29.68} (0.6) & 26.63 (1.3) & \textbf{62.67} (1.7) & 36.32 (1.0) & \textbf{53.42} (0.5)\\
\noalign{\vskip\aboverulesep}\cdashline{2-12}[2pt/1.2pt]\noalign{\vskip\belowrulesep}

& MHF (H3B10K) & 16.67 (0.5) & 44.84 (0.9) & 58.33 (4.5) & 26.40 (2.0) & \textbf{60.87} (1.0) & 27.56 (0.4)	& 27.27 (1.4) & 56.03 (1.6)	& \textbf{37.42} (1.0) & 48.61 (0.6)\\

& MHF (H4B16K) & 18.80 (0.4) & \textbf{45.61} (1.0) & 58.67 (4.5) & 26.13 (2.0) & 59.92 (1.1) & 27.81 (0.4)	& \textbf{28.52} (1.4) & 58.23 (1.7) & 36.15 (1.0) & 50.69 (0.6)\\
\bottomrule

\end{NiceTabular}
}
\caption{Performance of standard and our \textsc{MultiHashFormer} models with standard deviations over three runs in parentheses. Bold denotes best performance.}
\label{table:performance_multi_run_100M}

\end{table*}

\section{Comparison to Token-Free Models}
Although tokenizer-free/byte-level LMs mentioned in \S\ref{sec:token-free} operate on a different scope than \textsc{MultiHashFormer}, we include the performance of T-FREE~\cite{deiseroth-etal-2024-free}, which embeds words through sparse activations over locality-sensitive hashed character trigrams, for reference. Since T-FREE does not provide publicly accessible pre-trained checkpoints, we reproduced T-FREE at the 1B parameter scale using white-space tokenization and the 100K most frequent tokens extracted from our training corpus, matching the T-FREE original settings. Due to limited computational resources, we trained this model on 20B tokens and compared it against the 20B-token checkpoints of our \textsc{MultiHashFormer} and Standard baselines. The results are detailed in Table~\ref{table:performance_1B_tfree}. \textsc{MultiHashFormer} consistently outperform T-FREE across all tasks.

\begin{table*}[!t]
\centering
\small
\resizebox{\linewidth}{!}{
\begin{NiceTabular}{lcccccccccc}
\toprule
&\textbf{Lang. Model.} & \multicolumn{5}{c}{\textbf{Commonsense Reasoning}} & \multicolumn{3}{c}{\textbf{Reading Comprehension}} & \textbf{CR+RC}\\
\cmidrule(lr){2-2}   
\cmidrule(lr){3-7}   
\cmidrule(lr){8-10}  
\cmidrule(lr){11-11} 

\textbf{Model} & \textbf{LAMBADA} & \textbf{ARC-E} & \textbf{COPA} & \textbf{OBQA} & \textbf{PIQA} & \textbf{HellaSwag} & \textbf{RACE} & \textbf{SciQ} & \textbf{SIQA} & \textbf{ReCoRD}\\
\midrule

Standard & 30.41 & 59.85 & 64.00 & 31.60 & 65.78 & 38.99 & 29.38 & 69.90 & 39.10 & 63.20\\
Standard+2L & 30.66	& 60.23	& 65.00	& 31.20	& 65.02	& 39.85 & 29.28 & 72.10	& 38.89	& 63.78\\
\noalign{\vskip\aboverulesep}\cdashline{1-11}[2pt/1.2pt]\noalign{\vskip\belowrulesep}

T-FREE & NA	& 27.61	& 55.00	& 32.80	& 59.47	& 27.65	& 22.78 & 26.90	& 33.98	& 45.36\\
\noalign{\vskip\aboverulesep}\cdashline{1-11}[2pt/1.2pt]\noalign{\vskip\belowrulesep}

MHF (H3B10K) & \textbf{35.78} & 61.95 & 65.00 & \textbf{34.00} & \textbf{68.17} & 40.45 & 29.00 & 68.30 & 40.48 & 64.51\\

MHF (H4B16K) & 35.34 & \textbf{64.35} & \textbf{71.00} & 29.80 & 66.49 & \textbf{41.32} & \textbf{31.39} & \textbf{70.20} & \textbf{41.66} & \textbf{64.90}\\

\bottomrule

\end{NiceTabular}
}
\caption{Performance of standard, T-FREE and our \textsc{MultiHashFormer} models pretrained for 12B tokens at 1B scale. Bold denotes best performance. LAMBADA is marked as NA for T-FREE due to the length divergence introduced by its white space tokenizer.}
\label{table:performance_1B_tfree}

\end{table*}

\section{Computational Inference Overhead During Probability Re-normalization}
Re-normalizing over the actual vocabulary requires an explicit summation over valid token combinations and scales linearly with the actual vocabulary size at inference time. However, on modern hardware, these calculations are highly parallelizable and execute with a negligible wall-clock compute overhead. We evaluated our MHF (H4B16K; 128M embedding parameters) without any sophisticated code optimization at the 1B parameter scale on the LAMBADA dataset, ensuring that the evaluation context length matched the sequence length used during pre-training. We perform inference across the entire dataset 10 times and report the average runtime. On one AMD MI300X GPU, MHF (H4B16K) takes 97.38 ms, whereas the Standard baseline takes 96.42 ms per sample on average, meaning MHF is effectively on par with the Standard baseline.

\end{document}